\documentclass[letterpaper]{article} 
\usepackage[draft]{aaai2026}  
\usepackage{times}  
\usepackage{helvet}  
\usepackage{courier}  
\usepackage[hyphens]{url}  
\usepackage{graphicx} 
\usepackage{natbib}  
\usepackage{caption} 
\usepackage{algorithm}
\usepackage{algorithmic}

\usepackage{booktabs}
\usepackage{pifont}
\newcommand{\cmark}{\ding{51}}
\newcommand{\xmark}{\ding{55}}

\usepackage{newfloat}
\usepackage{listings}
\DeclareCaptionStyle{ruled}{labelfont=normalfont,labelsep=colon,strut=off} 
\floatstyle{ruled}
\newfloat{listing}{tb}{lst}{}
\floatname{listing}{Listing}
\title{FinGAIA: A Chinese Benchmark for AI Agents in Real-World Financial Domain}
\author {
    Lingfeng Zeng\textsuperscript{\rm 1},
    Fangqi Lou\textsuperscript{\rm 1},
    Zixuan Wang\textsuperscript{\rm 1},
    Jiajie Xu\textsuperscript{\rm 1},
    Jinyi Niu\textsuperscript{\rm 2},
    Mengping Li\textsuperscript{\rm 1},\\
    Yifan Dong\textsuperscript{\rm 1},
    Qi Qi\textsuperscript{\rm 1},
    Wei Zhang\textsuperscript{\rm 1},
    Ziwei Yang\textsuperscript{\rm 1},
    Jun Han\textsuperscript{\rm 1},
    Ruilun Feng\textsuperscript{\rm 1},\\
    Ruiqi Hu\textsuperscript{\rm 1},
    Lejie Zhang\textsuperscript{\rm 1},
    Zhengbo Feng\textsuperscript{\rm 1},
    Yicheng Ren\textsuperscript{\rm 1},
    Xin Guo\textsuperscript{\rm 1},
    Zhaowei Liu\textsuperscript{\rm 1},\\
    Dongpo Cheng\textsuperscript{\rm 1},
    Weige Cai\textsuperscript{\rm 1},
    Liwen Zhang\textsuperscript{\rm 1}\thanks{Corresponding Author.}
}
\affiliations {
    \textsuperscript{\rm 1}Shanghai University of Finance and Economics\\
    \textsuperscript{\rm 2}Fudan University\\
    {zhang.li}wen@shufe.edu.cn 
}

\usepackage{multirow}

\begin{document}

\maketitle

\begin{abstract}
The booming development of AI agents presents unprecedented opportunities for automating complex tasks across various domains. However, their multi-step, multi-tool collaboration capabilities in the financial sector remain underexplored. This paper introduces FinGAIA, an end-to-end benchmark designed to evaluate the practical abilities of AI agents in the financial domain. FinGAIA comprises 407 meticulously crafted tasks, spanning seven major financial sub-domains: securities, funds, banking, insurance, futures, trusts, and asset management. These tasks are organized into three hierarchical levels of scenario depth: basic business analysis, asset decision support, and strategic risk management. We evaluated 10 mainstream AI agents in a zero-shot setting. The best-performing agent, ChatGPT, achieved an overall accuracy of 48.9\%, which, while superior to non-professionals, still lags financial experts by over 35 percentage points. Error analysis has revealed five recurring failure patterns: Cross-modal Alignment Deficiency, Financial Terminological Bias, Operational Process Awareness Barrier, among others. These patterns point to crucial directions for future research. Our work provides the first agent benchmark closely related to the financial domain, aiming to objectively assess and promote the development of agents in this crucial field. Partial data is available at \url{https://github.com/SUFE-AIFLM-Lab/FinGAIA}

\end{abstract}

\begin{figure*}[!h]   
    \centering
    \includegraphics[width=\textwidth]{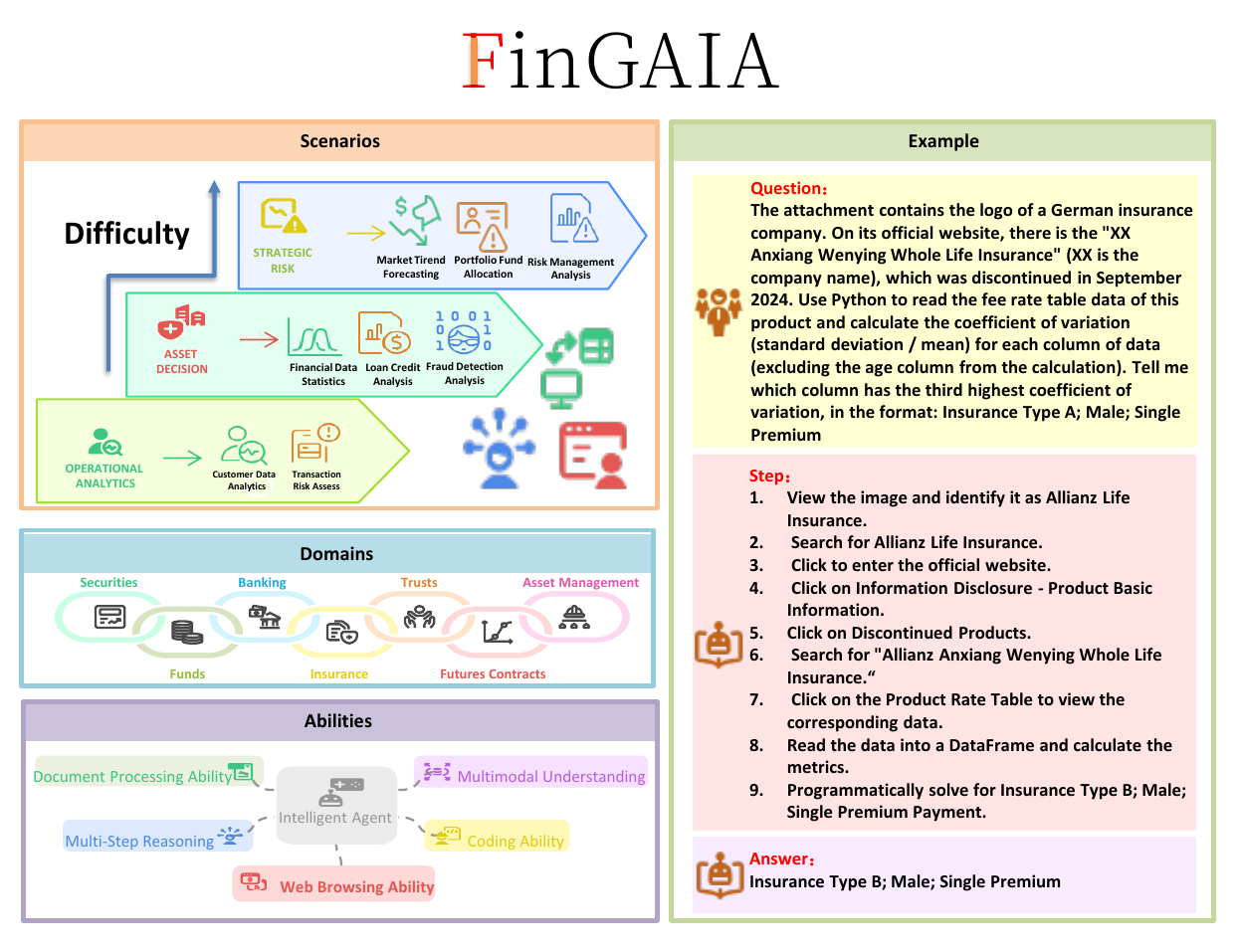}
    \caption{FinGAIA offers an AI Agent evaluation framework for full-process financial operations. From the perspective of business depth, it features three major scenarios: Operational Analytics, Asset Decision, and Strategic Risk. Each primary scenario includes 2, 3, and 3 sub-scenarios, respectively, accurately reflecting the diverse business needs of the financial industry. The upper part of the image shows the 7 common financial sub-domains covered by FinGAIA's constructed questions. The left section outlines FinGAIA's overall structure, where business depth increases as the color gradient darkens. Simultaneously, the framework progressively raises demands on the model's understanding and analytical capabilities regarding financial operations.The lower left part details the specific domains and competencies required for the AI Agent to solve problems, while the right panel provides a concrete example, including the question, steps, and answer.}
    \label{fig:frame}
\end{figure*}

%

\begin{table*}[ht]
\centering
\begin{tabular}{l c c c c c c}
    \toprule[2pt]
    \textbf{Benchmarks} & \textbf{Question Type} & \textbf{\shortstack{Multi-level\\Difficulty}} & \textbf{\shortstack{Scenario\\Depth}} & \textbf{\shortstack{Multi-tool\\Support}} & \textbf{\shortstack{Dynamic\\Environment}} & \textbf{\shortstack{Financial\\Domains}} \\
    \midrule[1pt]
    ToolBench & MC & \xmark & \xmark & \cmark & \xmark & \xmark \\
    API-Bank & OE & \xmark & \xmark & \xmark & \xmark & \xmark \\
    WebArena & MC & \xmark & \cmark & \xmark & \cmark & \xmark \\
    AgentBench & MC & \xmark & \cmark & \xmark & \cmark & \xmark \\
    GAIA & MC+OE & \cmark & \cmark & \cmark & \cmark & \xmark \\
    \textbf{FinGAIA (ours)} & MC+OE & \cmark & \cmark & \cmark & \cmark & \cmark \\
    \bottomrule[2pt]
\end{tabular}
\caption[Agent Benchmarks Comparison]{Comparison of Agent Benchmarks. Question Type abbreviations: MC (Multiple-Choice questions), OE (Open-Ended questions).}
\label{tab:benchmarks}
\end{table*}

\section{Introduction}
The rapid evolution of AI agents is revolutionizing how various industries tackle complex and intricate problems~\cite{yao2023react,sumers2023cognitive,ahn2022can,schick2023toolformer}. Their application mode, which involves autonomously planning and achieving goals based on large language models~\cite{zhang2409xlam,li2023metaagents,rafailov2023direct,jing2019task,nijkamp2022codegen}, has garnered widespread public attention. The empowering role of agents is particularly significant in the financial sector, where they are crucial for automating the processing of vast amounts of data, analyzing market trends, and optimizing decision-making processes. This ultimately helps financial institutions enhance efficiency and mitigate risks. Such integration necessitates that agents possess a comprehensive suite of core capabilities, including a deep understanding of industry knowledge, proficient tool utilization, and integrated reasoning abilities to handle complex tasks. However, due to the stringent requirements and specialized nature of the financial industry, effectively evaluating the true capabilities of AI agents presents unique challenges.

While researchers have gradually introduced specialized benchmarks for the financial domain, such as FinEval~\cite{zhang2023fineval}, FinQA~\cite{chen2021finqa}, CFLUE~\cite{zhu2024benchmarking}, CFinBench~\cite{nie2024cfinbench} and MME-Finance ~\cite{gan2024mme}, these benchmarks remain predominantly text-centric or image-centric. They generally lack systematic testing for multi-file, multi-tool collaboration, and multi-step execution flows, which are critical capabilities required for agents to empower the financial sector. Similarly, most existing AI benchmarks exhibit significant shortcomings when evaluating agents' capabilities in the financial domain. Common general-purpose benchmarks often focus on text-based question answering, function synthesis, or tool-use tasks~\cite{starace2025paperbench,zheng2025mcu,kokane2025toolscan,yao2022webshop,jimenez2023swe,ruan2023identifying,kokane2024spectool,chang2024agentboard,liu2024agentlite,yao2025tau}. Although effective in measuring a model's language understanding and coding abilities, they fail to meet the assessment demands of complex business processes within financial scenarios.

To address the gaps in current work and advance the development of financial AI agents, we propose FinGAIA the first end-to-end agent evaluation benchmark specifically tailored for financial scenarios. FinGAIA systematically integrates three core dimensions: industry knowledge, tool utilization, and task complexity, covering the entire financial workflow from basic information retrieval and multi-modal document analysis to code-based computation and the coordination of multi-tool, multi-step decision-making. FinGAIA comprises a total of 407 tasks, encompassing seven major financial sub-domains: securities, funds, banking, insurance, futures, trusts, and asset management. These tasks are designed based on extensive discussions with financial domain experts and constructed using real-world financial data, undergoing professional screening and structuring to ensure their authenticity, relevance, and evaluation utility. We adopt a hierarchical evaluation framework, categorizing tasks into basic business analysis, asset decision support, and strategic risk management, corresponding to different cognitive complexities and operational requirements. In our experimental evaluation, we tested 10 mainstream AI Agent frameworks, including both closed-source and locally deployed open-source agents. The evaluation employed a strict "zero-shot prompting" paradigm, supplemented by manual review and LLM-as-Judge for result verification.

In summary, our work makes the following contributions:
\begin{itemize}
  \item \textbf{The First AI Agent Benchmark in the Financial Domain} We constructed FinGAIA, which includes 407 meticulously designed tasks based on real-world scenario data, covering seven major financial sub-domains and three levels of scenario depth. This fills the evaluation gap for agents in the financial sector, providing a comprehensive and in-depth assessment system for evaluating agents' financial capabilities.  
  \item \textbf{High-Quality and Practice-Oriented Dataset} All tasks in FinGAIA were formulated through discussions with financial experts, and the entire question creation process was executed manually. On average, each question required approximately 90 minutes to complete the full design, annotation, and verification process from extracting data from real financial scenarios. This rigorous construction process and quality control ensure the dataset's authority while providing a practical, workflow-oriented evaluation framework for eight major scenarios.  
  \item \textbf{Multi-Level Comparative Analysis} Our work includes a comparison of agent capabilities against those of ordinary financial undergraduate students and Ph.D. students in finance, offering valuable insights into agents' abilities in professional domains. The results indicate that agents still have room for improvement in specialized areas.
\end{itemize}

\section{Related Work}\label{sec:s2}

\subsection{Agent Development History}
In recent years, Agent technology has evolved from basic multi-step tool invocation to autonomous end-to-end systems. Initially, frameworks like LangChain proposed modular multi-step reasoning and external tool chaining mechanisms~\cite{chase2023langchain}. Subsequently, Toolformer enabled Agents to autonomously determine and invoke appropriate tools through self-supervision, significantly enhancing decision-making efficiency~\cite{schick2023toolformer}. ReAct integrated reasoning with action decision-making, enabling real-time API calls and feedback~\cite{yao2023react}. Building on this, open-source projects like AutoGPT and BabyAGI introduced a "decompose-execute-feedback" loop, achieving end-to-end adaptive optimization~\cite{significant-gravitas-2023-autogpt,nakajima:babyagi}. Concurrently, OpenAI's Function Calling and Plugins ecosystem standardized interfaces for secure and efficient interaction between Agents and third-party services~\cite{openai:functioncalling,openai:chatgptplugins}. Academia has also provided solutions in multi-model collaboration, task planning, and operational monitoring, offering systematic support for Agent development and operation~\cite{shen2023hugginggpt,prasad2023adapt,smith:langsmith}. In the financial domain, Agents have undergone rapid iteration: from their initial form as digital advisors (Robo-advisors), providing algorithmic asset allocation for individual investors with low-cost, automated portfolio management services~\cite{fisch2019emergence}; to domain-specific pre-training systems for financial corpora~\cite{yang2020finbert}; and further to integrated real-time market data capture, model inference, and interpretable report generation, driving the intelligent automation of financial decision-making processes~\cite{huang2024open}.

\subsection{Agent Benchmark History}
The advent of agentic architectures—capable of multi-step reasoning, dynamic tool chaining, and multimodal interaction—has outstripped the capabilities of traditional benchmarks built for static QA or simple API calls. Early platforms such as ToolBench~\cite{qin2023toolllm} and API-Bank~\cite{li2023api} validate a model’s proficiency in predefined API invocations but rely on highly structured, closed environments that may encourage pattern memorization over genuine reasoning. WebArena~\cite{zhou2023webarena} and AgentBench~\cite{liu2023agentbench} evaluate robustness in simulated web and system-interaction scenarios yet depend heavily on familiarity with specific DOM structures or command-line syntaxes. GAIA~\cite{mialon2023gaia} represents a significant advance by driving agents to interact end-to-end with live web tools—encompassing web browsing, document parsing, and code execution—to assess tool collaboration, long-chain reasoning, and self-correction. However, GAIA’s scope is largely confined to general web and document operations and does not address finance-specific requirements such as regulatory review, real-time market data retrieval, multimodal report parsing, or proprietary system API integration.

In order to bridge these gaps, we propose FinGAIA, the first end-to-end agent evaluation benchmark explicitly tailored to financial scenarios. FinGAIA systematically integrates three core dimensions—industry knowledge, tool utilization, and task complexity—and spans the full financial workflow: from basic information retrieval and multimodal document analysis to Python-based computation and orchestrated multi-tool, multi-step decision making. By capturing the intricacies of real-world financial processes, FinGAIA effectively addresses the limitations of existing benchmarks and provides a comprehensive framework for assessing agentic performance in professional financial contexts.

\section{FinGAIA}
\label{section:3}
\subsection{Overview}

We propose FinGAIA, the first benchmark specifically designed to evaluate the capabilities of intelligent agents within the financial domain. FinGAIA aims to systematically measure agents’ task execution performance and professional competence in realistic financial workflows. Centered on highly faithful simulations of real-world financial scenarios, the benchmark comprehensively spans tasks ranging from basic business comprehension to complex strategic decision-making. It focuses on assessing agents’ abilities in financial language understanding, data integration and analysis, tool-based collaboration, and multi-step reasoning. The overall architecture is illustrated in Figure~\ref{fig:frame}.
In terms of data generation and quality assurance, FinGAIA follows a rigorous construction pipeline. Four finance professors designed the task scenarios and authored the initial questions. These were annotated by six professionally trained undergraduate students with a solid background in finance, and then reviewed across multiple dimensions by four domain experts. The full dataset comprises 407 questions, covering a wide range of subdomains including securities, funds, and banking. These tasks are designed to reflect typical user demands in asset allocation, policy interpretation, risk assessment, and market analysis.

FinGAIA establishes a three-tiered capability evaluation framework, encompassing three difficulty levels and eight representative financial business scenarios. Each level corresponds to a different cognitive requirement:  

\noindent\textbf{Level 1} – Basic Business Analysis: evaluates agents’ understanding of fundamental financial knowledge and their ability to process multimodal financial information;  

\noindent\textbf{Level 2} – Asset Decision Support: focuses on tasks involving moderate complexity, requiring information integration, logical reasoning, and flexible tool use;  

\noindent\textbf{Level 3} – Strategic Risk Management: targets highly complex tasks requiring multi-tool coordination and strategic planning, to comprehensively assess reasoning ability and domain expertise under realistic financial conditions.

Overall, FinGAIA provides a structured, diverse, and contextually grounded benchmark for evaluating intelligent agents in the financial vertical, enabling comprehensive and objective assessment of their professional competencies, mainly including web browsing capability, document processing capability, multimodal understanding, coding ability, and computational ability.
The distribution of tasks across business scenarios and dataset statistics is shown in Appendix Table~\ref{tab:fingaia_dist}, and representative task examples are included in the Appendix A.

\subsection{Question Generation and Quality Control}

The FinGAIA dataset encompasses data derived from a wide range of real-world financial scenarios. Major data sources include legal and regulatory text, publicly available market transaction data, financial news and analytical reports, and key indicators of financial products. All real-world data used in the benchmark have been carefully verified to ensure they are free of copyright restrictions, with explicit source attribution provided in each task.

During the task construction process, to more accurately simulate the real-world application scenarios of intelligent agents in the financial domain, four professors with financial expertise designed eight representative financial business scenarios distributed across three progressively difficult levels. Tasks were manually created under each scenario to reflect the corresponding complexity. The resulting tasks were annotated by six professionally trained undergraduate students majoring in finance, all of whom possessed a solid foundation in financial knowledge. The training program covered key aspects such as verifying whether each task aligned with its designated difficulty level and scenario, annotating the correct answer, estimating task difficulty, identifying the tools involved, and assessing the time required for completion. Large-scale annotation was only conducted after the annotators passed a qualification test and evaluation, ensuring both consistency and high quality in the labeling process.

In the expert review stage, each task was thoroughly inspected by four industry professionals, each with over a decade of financial experience. Review criteria included alignment with real industry practice, logical soundness of the task structure, regulatory compliance, precision of financial terminology, and contextual relevance to the business scenario. A task was accepted into the dataset only if unanimously approved by all four experts, ensuring that each question was well-designed, unambiguous, and reflective of authentic financial operations.
Through this rigorous data generation and quality control pipeline, we constructed FinGAIA, a high-quality benchmark for evaluating financial AI agents. The final dataset includes 407 expert-validated tasks. 


\begin{table*}[t] 
\centering
\setlength{\tabcolsep}{4pt} 
\begin{tabular}{lccccccccc}
\toprule[2pt]
\multirow{2}{*}{\textbf{Agent}} & \multicolumn{2}{c}{\textbf{Operational Analytics}} & \multicolumn{3}{c}{\textbf{Asset Decision}} & \multicolumn{3}{c}{\textbf{Strategic Risk}} & \multirow{2}{*}{\textbf{WA}} \\
\cmidrule(lr){2-3} \cmidrule(lr){4-6} \cmidrule(lr){7-9}
& \textbf{CDA} & \textbf{TRA} & \textbf{FDS} & \textbf{LCA} & \textbf{FRA} & \textbf{RMA} & \textbf{PFA} & \textbf{MTF} & \\
\midrule[1pt]
ChatGPT(DeepResearch) & \textbf{44.7} & \textbf{57.1} & \textbf{50.5} & \textbf{58.1} & \textbf{39.0} & \textbf{57.1} & \textbf{47.5} & \textbf{37.3} & \textbf{48.9} \\
Perplexity DeepResearch & 31.9 & 52.4 & 30.7 & 48.8 & 26.8 & 42.9 & 35.0 & 27.5 & 37.0 \\
Cashcat DeepResearch & 29.8 & 52.4 & 21.8 & 48.8 & 31.7 & 26.2 & 35.0 & 19.6 & 33.2 \\
Kimi & 25.5 & 38.1 & 23.8 & 46.5 & 24.4 & 35.7 & 17.5 & 17.7 & 28.6 \\
Gemini-2.5-pro(DeepResearch) & 34.0 & 38.1 & 26.7 & 39.5 & 12.2 & 38.1 & 25.0 & 11.8 & 28.2 \\
Grok DeeperSearch & 23.4 & 35.7 & 15.8 & 34.9 & 2.4 & 31.0 & 35.0 & 19.6 & 24.7 \\
Reportify & 14.9 & 26.2 & 25.7 & 30.2 & 26.8 & 26.2 & 20.0 & 19.6 & 23.7 \\
OWL & 19.2 & 31.0 & 5.9 & 34.9 & 14.6 & 31.0 & 22.5 & 15.7 & 21.8 \\
Kompas AI & 25.5 & 9.5 & 19.8 & 20.9 & 14.6 & 16.7 & 22.5 & 5.9 & 16.9 \\
AutoGLM & 6.4 & 11.9 & 11.9 & 14.0 & 12.2 & 16.7 & 20.0 & 11.8 & 13.1 \\
\midrule
\textbf{Avg. Score} & 25.5 & 35.2 & 23.2 & 37.7 & 20.5 & 32.2 & 28.0 & 18.7 & -- \\
Human Expert & 82.5 & 88.3 & 85.1 &90.2& 79.6 & 92.7 & 83.7 & 75.8 & 84.7 \\
\bottomrule[2pt]
\end{tabular}
\caption{Main Results. The higher the value in the table, the higher the accuracy of the financial AI model. The evaluation assesses performance across three capability tiers: Operational Analytics covering Customer Data Analytics(CDA) and Transaction Risk Assessment(TRA); Asset Decision comprising Financial Data Statistics(FDS), Loan Credit Analysis(LCA), and Fraud Detection Analysis(FDA); and Strategic Risk evaluating Risk Management Analysis(RMA), Portfolio Fund Allocation(PFA), and Market Trend Forecasting(MTF), concluding with the calculation of the Weighted Average (WA) score for each model. The highest score in each scenario is bolded in the table. }
\label{tab:financial-results}
\end{table*}

\subsection{FinGAIA Question Architecture}

The FinGAIA benchmark is built upon representative financial business scenarios and establishes a three-tiered evaluation framework designed to systematically assess the overall performance of intelligent agents in realistic financial environments. The benchmark consists of 407 high-quality tasks, each rigorously validated by experts. Tasks are categorized into three ascending difficulty levels based on task complexity and the depth of required capabilities: Level 1 (Basic Business Analysis), Level 2 (Asset Decision Support), and Level 3 (Strategic Risk Management). These three levels respectively cover 2, 3, and 3 distinct financial scenarios. As the difficulty level increases, both the types of questions and the complexity of tasks intensify, enabling a comprehensive evaluation of an agent's abilities in financial language understanding, information integration, tool utilization, and complex decision reasoning. Representative examples of each level are provided in the Appendix A. The average number of tool invocations per scenario are provided in Table~\ref{tab:fingaia_dist}.

\noindent\textbf{Level 1: Basic Business Analysis} This level primarily evaluates an agent’s grasp of fundamental financial knowledge and its ability to quickly process multi-modal information. Tasks at this level are structurally simple, typically requiring no more than five steps and the use of only one or two tools. The focus is on reading, interpreting, and performing basic calculations over essential financial inputs such as customer data and transaction charts. Suitable for entry-level tasks like customer service or preliminary transaction screening, Level 1 includes two representative scenarios: Customer Data Analytics and Transaction Risk Assessment, comprising a total of 89 tasks.

\noindent\textbf{Level 2: Asset Decision Support} Level 2 focuses on medium-complexity analysis and decision-making tasks, requiring agents to demonstrate stronger information synthesis and strategy formulation capabilities. Tasks typically involve multiple data sources, increased reasoning steps from 5 to 7, and the integration of more than two tools. This level emphasizes logical reasoning, flexibility in tool usage, and deeper domain understanding, making it suitable for mid-level financial operations such as loan evaluation, asset assessment, and fraud detection. It covers three scenarios: Loan Credit Analysis, Financial Data Statistics, and Fraud Detection Analysis, with a total of 185 tasks.

\noindent\textbf{Level 3: Strategic Risk Management} Designed for high-stakes decision-making and advanced risk management, Level 3 features tasks that closely reflect real-world financial practices. These tasks involve a greater number of steps around 10 and require coordinated use of multiple tools, including sequential tool invocation and parameter tuning. Agents are expected to exhibit precise semantic comprehension and integration of financial expertise. This level comprehensively evaluates the agent's peak performance under complex financial conditions. It includes three core business scenarios: Portfolio Fund Allocation, Risk Management Analysis, and Market Trend Forecasting, comprising 133 tasks.

By constructing a tiered evaluation system grounded in realistic financial workflows, FinGAIA offers a scientific, comprehensive, and practically relevant benchmark for assessing the capabilities of financial AI agents.

\begin{table*}[t] 
\centering 
\begin{tabular}{lccccc}
\toprule[2pt]
\textbf{Source} & \textbf{Category} & \textbf{OA} & \textbf{AD} & \textbf{SR} & \textbf{Average} \\
\midrule[1pt]
Human & Non-experts & 58.3 & 47.1 & 35.2 & \textbf{46.9} \\
& Experts & 87.6 & 82.4 & 76.5 & \textbf{82.2} \\
\midrule
Closed-Source & ChatGPT (DeepResearch) & 68.2 & 52.4 & 28.2 & \textbf{49.6} \\
& Perplexity DeepResearch & 51.7 & 40.3 & 22.9 & \textbf{38.3} \\
\midrule
Open-Source & OWL & 25.5 & 20.8 & 15.2 & \textbf{20.4} \\
\bottomrule[2pt]
\end{tabular}
\caption{Performance comparison across non‐experts, experts and Agents. OA refers to Operational Analytics. AD means Asset Decision. SR represents Strategic Risk.}
\label{tab:Comparative Analysis table}
\end{table*}

\section{Experiment}
\label{s4}
\subsection{Agents}
We evaluated 10 currently mainstream Agent frameworks, consisting of 9 closed-source agents and 1 locally deployed open-source framework OWL \cite{owl-2025}. 9 closed-source agents were tested via their web interfaces, including ChatGPT DeepResearch \cite{chatgpt-deepresearch-2025}, Perplexity DeepResearch \cite{perplexity-2025}, Kimi \cite{kimi-2023}, Cashcat DeepResearch \cite{cashcat-2025}, Grok DeeperSearch \cite{grok-2025}, Gemini-2.5-pro \cite{gemini-2.5-pro-2025}, Reportify \cite{reportify-2024}, and Kompas AI \cite{kompasai-2024}. And AutoGLM \cite{autoglm-2024} was integrated via a browser plugin. Detailed information is provided in Appendix B.

\begin{figure*}[!h]   
    \centering
    \includegraphics[width=0.8\textwidth]{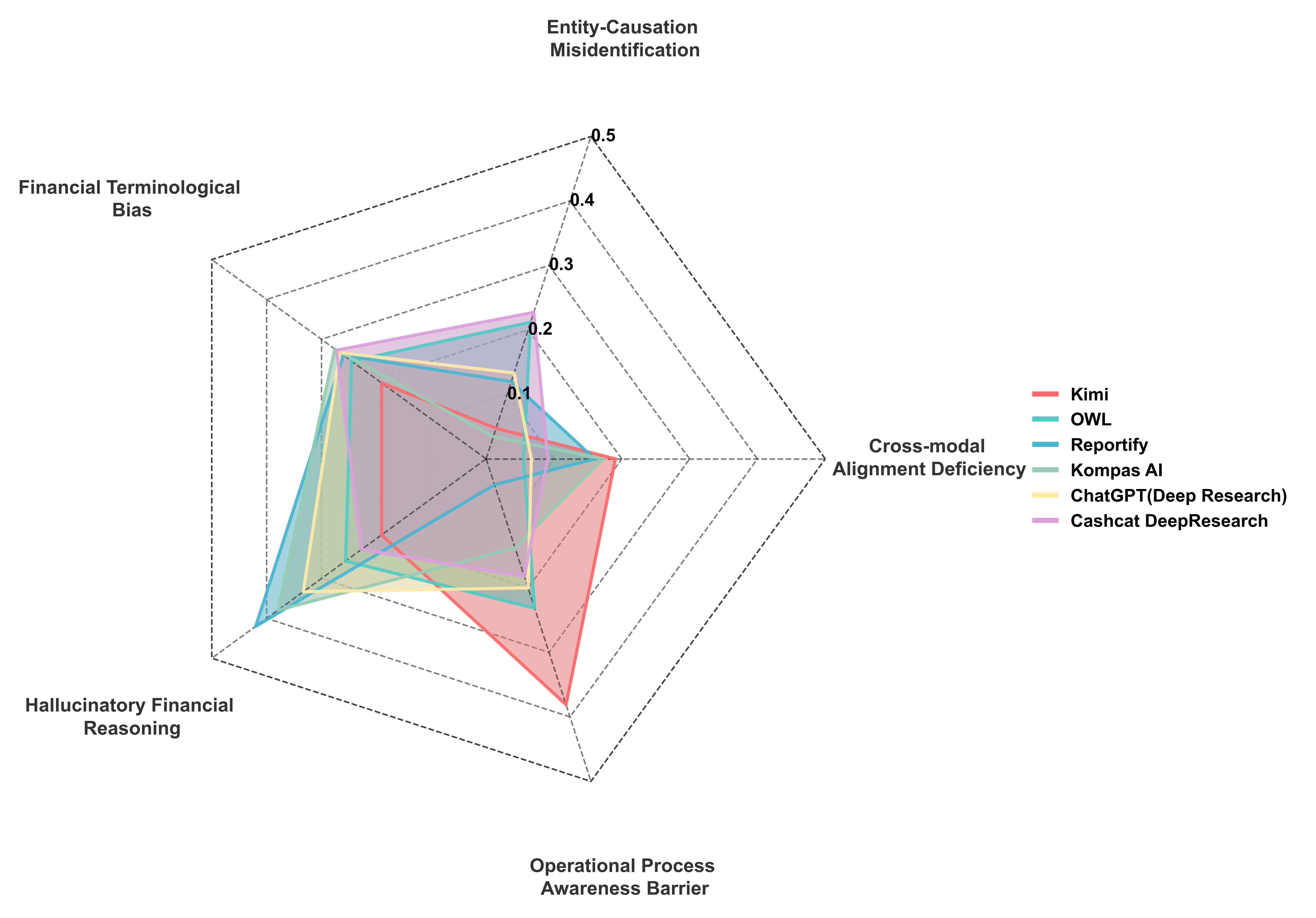}
    \caption{This radar chart compares error distributions across six financial AI agents in FinGAIA evaluations, with vertex position reflecting severity in five error types. Axis scales (0–0.5) quantify performance deviations for targeted improvement.}
    \label{erroranalysis}
\end{figure*}







\subsection{Evaluation Methods}
In the actual evaluation process, we employed differentiated testing protocols tailored to each agent’s tool‑invocation capabilities. For closed‑source agents accessed via web or application interfaces, we conducted manual zero‑shot interaction tests—providing only the task prompt and necessary attachments (e.g., Excel files, PDF reports) without any examples or guidance. For API‑enabled frameworks, we submitted all test items in batch through a unified API endpoint, embedding a predefined system prompt to elicit standardized, structured outputs.

During the evaluation, we recorded each agent’s score for correctly executed tasks across all test scenarios. To ensure the accuracy and consistency of the output, we primarily relied on manual review, supplemented by LLM-as-Judge's automatic interpretation~\cite{xu2023mmbench}, to perform item-by-item validation of the system's output.

After completing all data collection and scoring, we computed each agent’s accuracy—both across the full set of tasks and within the Level 1–3 stratified tasks—by dividing its total number of correct responses by the total number of questions. This method ensures a fair and reproducible comparison of each framework’s end-to-end performance and tool-coordination capabilities in financial scenarios.In Appendix B, we excluded tasks unprocessable due to unsupported file formats and recalculated accuracy on the remaining successful tasks to illustrate each agent’s performance in executable scenarios.

\section{Results}
\label{s5}
\subsection{Main Results}

%

We evaluated 10 mainstream Agents, as shown in Table~\ref{tab:financial-results}. Due to certain models' inability to read specific file types, some questions could not be effectively answered by certain models. Therefore, we excluded the results of models that could not be evaluated from the final assessment. Overall, the results showed no significant changes. For detailed information, please refer to Appendix Table \ref{tab:financial-results-in-assessable-scenarios}.  

The performance of financial AI agents varies significantly across different scenarios, demonstrating clear stratification. ChatGPT (DeepResearch) ranked first with an average weighted score (WA) of 48.9, exhibiting exceptional cross-level stability, particularly excelling in Risk Management Analysis (RMA 60.0) within strategic planning scenarios and Loan Credit Analysis (LCA 58.1) in tactical advisory scenarios. The second tier, consisting of Perplexity DeepResearch (WA 37) and Cashcat DeepResearch(WA 33.2), achieved an average score of 52.4 in Transaction Risk Assessment (TRA) for basic operational tasks, significantly exceeding the industry average by 17.2 points.  

Key scenario analysis reveals that market forecasting and fraud detection remain common industry weaknesses, with only ChatGPT demonstrating breakthrough performance in fraud detection (FRA 39.0). Further analysis indicates that the poorest-performing model, Grok DeeperSearch (FRA 2.4), suffers from severe cross-modal alignment deficiencies and anomaly pattern recognition blind spots.  

Although ChatGPT (DeepResearch) achieved the highest accuracy in this evaluation, this score remains significantly lower than that of human experts, particularly in Strategic Risk tasks, where agent performance was generally weak. This highlights the current limitations of large models in complex reasoning and deep comprehension tasks. It also suggests that while existing agents have made considerable progress in simpler or moderately complex problems, their performance in more challenging high-difficulty tasks still has substantial room for improvement. Future research and development must further enhance these models' reasoning capabilities, depth of problem-solving, and understanding of complex scenarios.

\subsection{Comparative Analysis}

To better compare agent capabilities and make substantive contributions to model research, we conducted a competition among agents, non-experts, and financial experts by stratifying and sampling 50 questions from the FinGAIA dataset. Considering differences in domain knowledge and to more accurately reflect the current development stage of agents, we selected the top-performing models from both open-source and closed-source categories for comparison. For human participants, we recruited undergraduate students without financial backgrounds to represent the non-expert group, while the role of financial experts was filled by PhD candidates specializing in finance. All participants were independent of any data annotation or evaluation processes related to this study, and all responses were completed without AI assistance.

As shown in Table~\ref{tab:Comparative Analysis table}, unlike Table~\ref{tab:financial-results}, we calculated the average results across three major scenarios and the overall score to compare human and model performance. The results demonstrate that the best-performing Agent models have surpassed non-experts in all three scenarios and in the composite average score. However, there remains a performance gap of over 32 between current agents and financial experts, indicating that agents still require continuous iterative optimization.

\subsection{Error Analysis}


For error analysis, we employed stratified sampling to select 50 of incorrect responses across all evaluated agents, systematically investigating deficiencies in agents' financial business capabilities. Through comprehensive error categorization, we identified six fundamental limitations in financial domain competence: Data Type Handling Error,  Financial Terminological Bias, Operational Process Awareness Barrier, Hallucinatory Financial Reasoning, and Entity-Causation Misidentification. These deficiencies collectively constrain model performance in specialized financial scenarios.

As evidenced by the error distribution analysis, most agents demonstrate relatively balanced error patterns. However, certain models (notably Kimi, Reportify, and Cashcat DeepResearch) exhibit statistically significant error concentrations in specific dimensions: Operational Process Awareness Barrier, Hallucinatory Financial Reasoning, and Entity-Causation Misidentification, respectively. These systematic biases result in disproportionately higher error rates in these categories compared to baseline expectations.Detailed methodological explanations, including sampling procedures and significance testing protocols, along with representative error case studies, are documented in Appendix B.

\section{Conclusion}
\label{s6}
This paper introduces FinGAIA, the first end-to-end benchmark meticulously designed to evaluate the capabilities of AI agents in the financial domain. The benchmark is structured around seven major financial sub-areas and organized into a three-tiered complexity hierarchy to systematically explore various agent capabilities.

Zero-shot evaluation results show that the best-performing model, ChatGPT (DeepResearch), achieved an overall accuracy of 48.9. While this performance surpasses that of human finance undergraduates, there remains a significant gap when compared to finance PhDs. This highlights a substantial room for improvement, especially in highly complex, multi-step tasks.
Furthermore, our comprehensive error analysis reveals five major capability deficiencies in agents when applied to real-world financial scenarios. As the first agent benchmark rooted in actual financial industry workflows, FinGAIA provides a structured, multi-level evaluation framework for measuring agents' end-to-end capabilities in financial practice.

We hope that the release of FinGAIA, along with its high-quality and diverse tasks, will advance agent research, foster a deeper understanding of real-world financial scenarios, and lay a solid foundation for the future development of more powerful and reliable financial intelligent agents. We plan to make the FinGAIA benchmark and its detailed evaluation results publicly available to encourage broader research community participation in exploring and enhancing financial agent capabilities.

\section*{Limitations}

Despite the significant progress FinGAIA has made in evaluating AI Agents in the financial domain, some limitations remain, requiring further research and development. Although certain FinGAIA tasks may inherently involve dynamic elements, there is a need for more explicit and in-depth exploration of time-series analysis, real-time market fluctuations, and rapidly evolving economic indicators. Secondly, the current evaluation primarily focuses on zero-shot performance. While this provides a baseline understanding of an agent's out-of-the-box capabilities, it does not fully capture the potential for AI Agents to adapt and improve with limited examples. Future research should further consider the potential for AI Agents to adapt to new or changing financial tasks and scenarios through few-shot learning. Lastly, although FinGAIA covers a wide range of financial sub-domains and business scenarios, different business scenarios have varying importance. Therefore, it is necessary to design more appropriate business scenario weights to truly evaluate the actual performance of AI Agents in the financial domain.

\bibliography{aaai2026}


\appendix
\section{A  Details of FinGAIA}
To promote transparency and facilitate the reproducibility of our findings, a portion of the FinGAIA benchmark data is provided. The data is hosted anonymously at: \url{https://github.com/SUFE-AIFLM-Lab/FinGAIA}

\subsection{Design and Examples of Financial Business Scenarios }
We detail the FinGAIA data in Table~\ref{tab:fingaia_dist}. We adopted a three-tiered structure for financial business scenarios, including basic business analysis, asset decision support, and strategic risk management. The detailed information for specific financial business scenarios is presented below.

Business Analysis includes the following two financial business scenarios:
\textbf{Customer Data Analysis} primarily involves reviewing customer information to understand their data. The core of this work is to identify patterns and insights that inform targeted marketing strategies, product development, and personalized customer experiences. Through this analysis, financial institutions can optimize their offerings and build stronger customer relationships. Below is an example task for Customer Data Analysis.
\textbf{Transaction Risk Assessment} focuses on identifying and evaluating potential risks associated with financial transactions. This includes analyzing the likelihood of default, fraud, or other financial losses. The objective of this work is to mitigate risks by implementing appropriate controls and making informed decisions on transaction approvals.

Asset Decision Support includes three core business scenarios:
\textbf{Financial Data Statistics} involves collecting, organizing, and interpreting numerical financial information. The main task is to derive meaningful insights and trends from large datasets using statistical methods. This work supports reporting, forecasting, and identifying anomalies within financial operations.
\textbf{Loan Credit Analysis} is the process of assessing the creditworthiness of loan applicants. It involves evaluating financial statements, credit history, and other relevant factors to determine the likelihood of repayment. The goal is to make informed lending decisions that balance risk and potential returns.
\textbf{Fraud Detection Analysis} focuses on identifying and preventing fraudulent activities within financial systems. This includes analyzing transaction patterns and anomalies to flag suspicious behavior. The goal of this work is to protect institutions and their clients from financial crimes and losses.

Strategic Risk Management includes three high-level financial task scenarios:
\textbf{Risk Management Analysis} is a broad discipline focused on identifying, assessing, and mitigating various financial risks. This work involves developing strategies and controls to minimize potential negative impacts on an institution's financial health.
\textbf{Portfolio Fund Allocation} involves strategically distributing capital among different assets within an investment portfolio. The primary task is to optimize the portfolio's risk-return profile based on investment objectives and market conditions. This ensures efficient capital utilization, thereby maximizing returns while managing acceptable levels of risk.
\textbf{Market Trend Forecasting} focuses on predicting future movements and directions in financial markets. This includes analyzing historical data, economic indicators, and various other factors to anticipate market changes. The goal of this work is to provide valuable insights for investment decisions, trading strategies, and risk management.

\begin{figure*}[!h]
   \centering
   \includegraphics[width=0.9\textwidth]{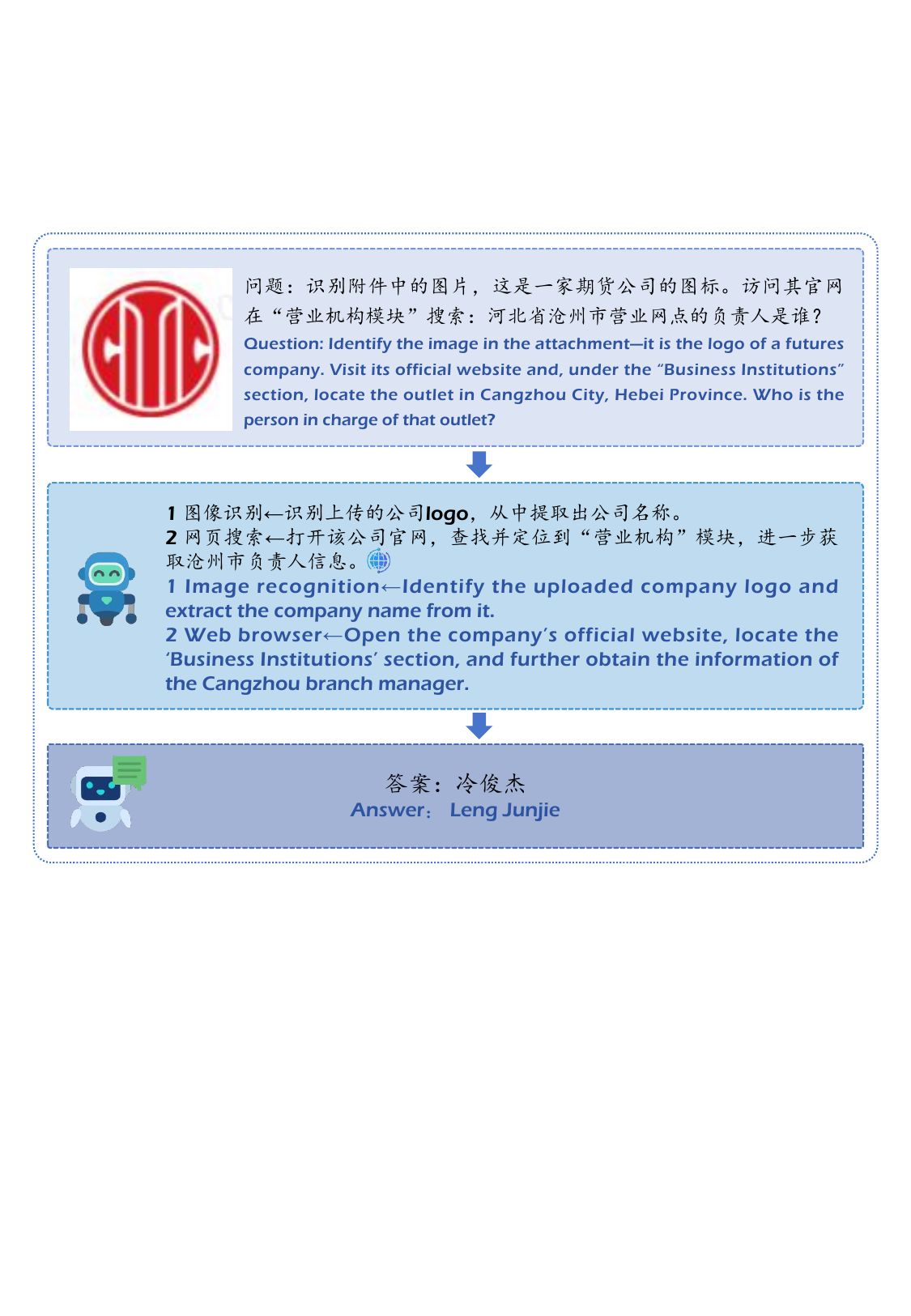}
   \vspace{-220pt}  
    \captionsetup{width=0.9\textwidth}
   \caption{This is a Customer Data Analytics scenario example focusing on branch manager information retrieval—the Agent must identify the futures company’s logo shown in the figure, navigate to its official website’s “Business Institutions” module, search for the Cangzhou outlet in Hebei Province, and locate the name of the person in charge to answer, “Who is the manager of the Cangzhou outlet?” This task tests the Agent’s operational analytics capabilities in logo recognition, website navigation, information retrieval, and rapid response, demonstrating the value of Operational Analytics Tier.}
   \label{Level1:CDA}
\end{figure*}

\begin{figure*}[!h]
   \centering
   \includegraphics[width=0.9\textwidth]{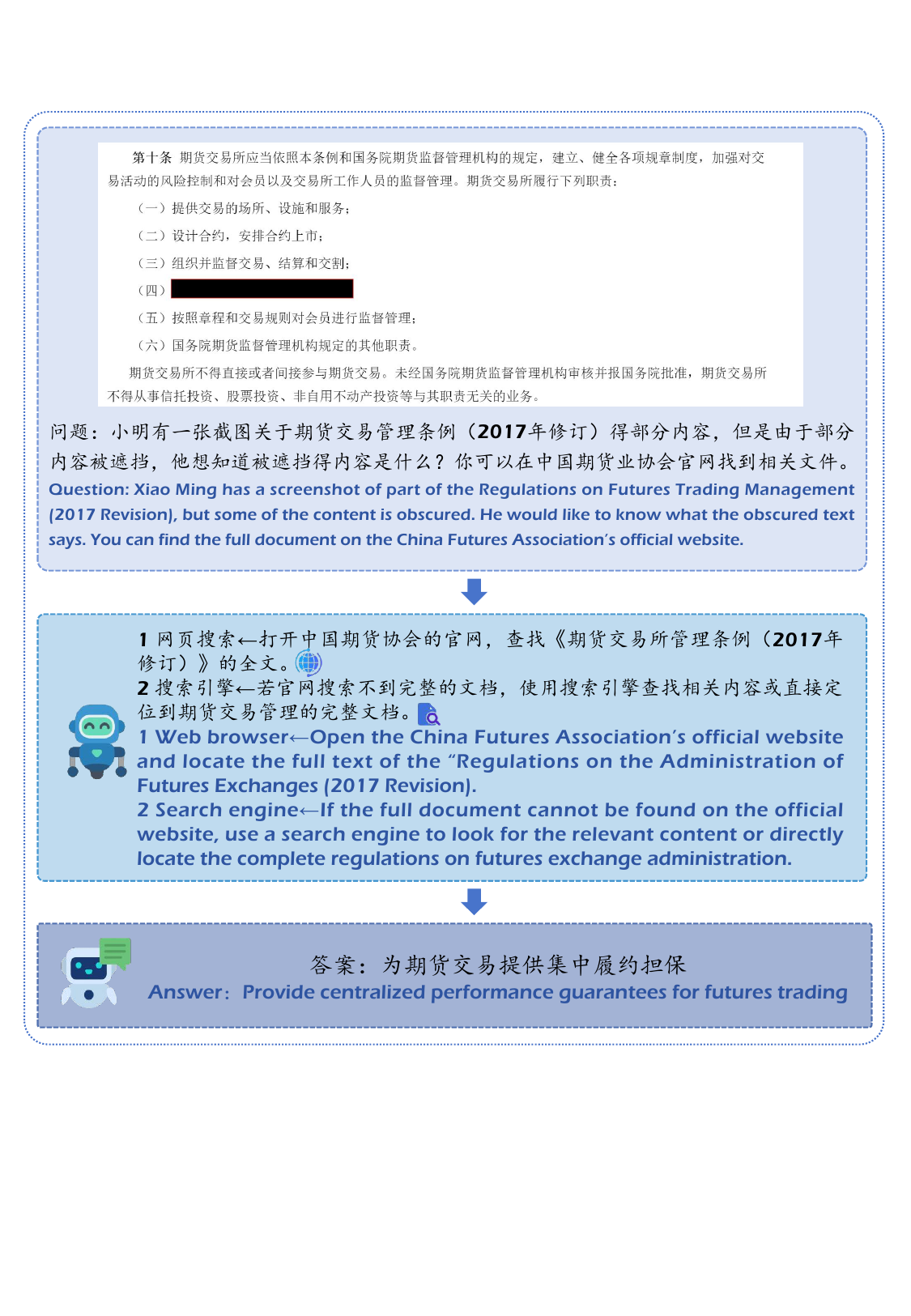}
    \vspace{-130pt}  
    \captionsetup{width=0.9\textwidth}
   \caption{This is a Transaction Risk Assessment scenario example focusing on a task to complete the text of Article 10, paragraph (4) of the Regulations on Futures Trading Management (2017 Revision)—the Agent must identify the screenshot’s source and clause, navigate to the China Futures Association’s “Regulations and Documents” section, download the full text, locate Article 10 (4), and extract the obscured content to answer the user’s question. This tests the Agent’s foundational business analysis capabilities in regulatory document retrieval, clause localization, and text extraction, demonstrating the value of Operational Analytics.}
   \label{Level1:TRA}
\end{figure*}

\begin{figure*}[!h]
   \centering
   \includegraphics[width=0.9\textwidth]{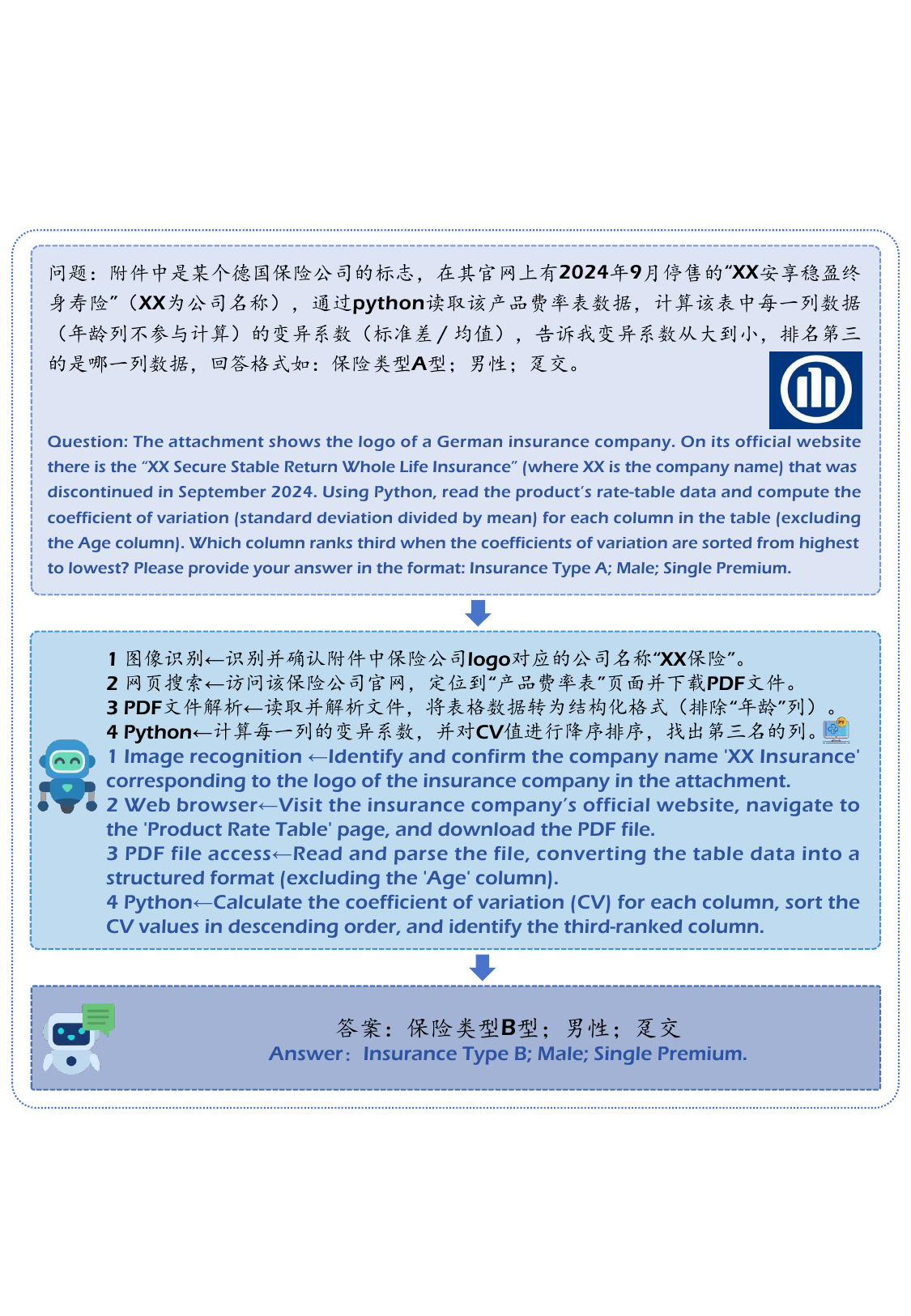}
    \vspace{-100pt}  
    \captionsetup{width=0.9\textwidth}
    \caption{This is a financial data statistics task. To answer this question, the Agent must first identify the German insurance company’s logo in the figure, then navigate to its official website to locate the rate table for the “XX Secure Stable Return Whole Life Insurance” product, use Python to read the table, compute the coefficient of variation for each column (excluding “Age”), sort these coefficients in descending order, and select the column that ranks third. This question tests the Agent’s ability to aggregate and parse financial data tables, calculate and rank basic statistical metrics (coefficient of variation), and carry out multi‐step information retrieval and programming execution, highlighting a rate sensitivity analysis scenario at the Asset Decision Tier.}
   \label{Level2:FDS}
\end{figure*}

\begin{figure*}[!h]
   \centering
   \includegraphics[width=0.9\textwidth]{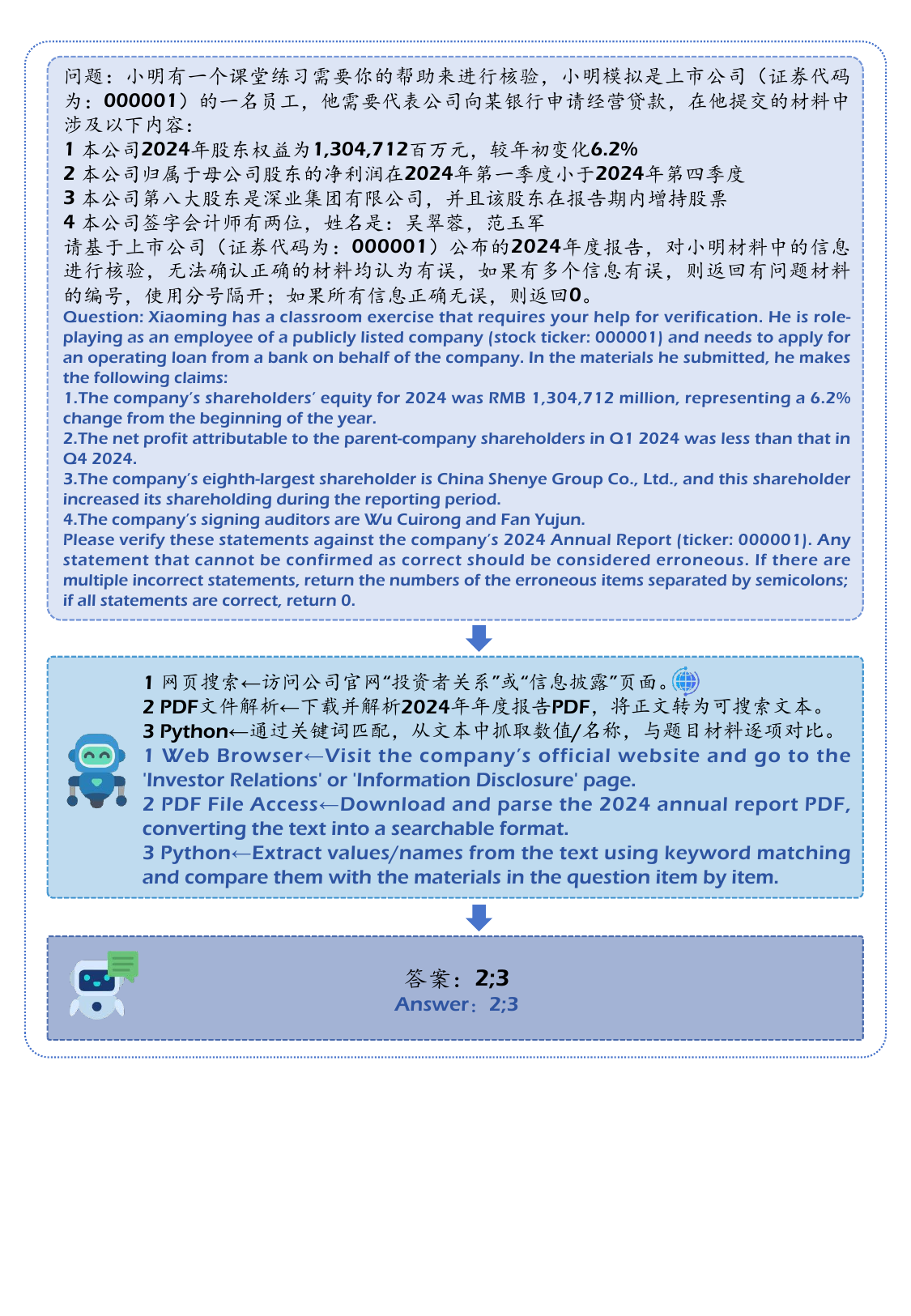}
    \vspace{-120pt}  
    \captionsetup{width=0.9\textwidth}
    \caption{A Loan Credit Analysis scenario example where the Agent consults the listed company’s 2024 Annual Report to verify four claims—total shareholders’ equity; Q1 vs. Q4 net profit; the identity and stake change of the eighth-largest shareholder; and the signing auditors’ names—and returns any incorrect item numbers, testing data extraction, comparative analysis, and text-matching skills at the Asset Decision Tier.}
   \label{Level2:LCA}
\end{figure*}

\begin{figure*}[!h]
   \centering
   \includegraphics[width=0.9\textwidth]{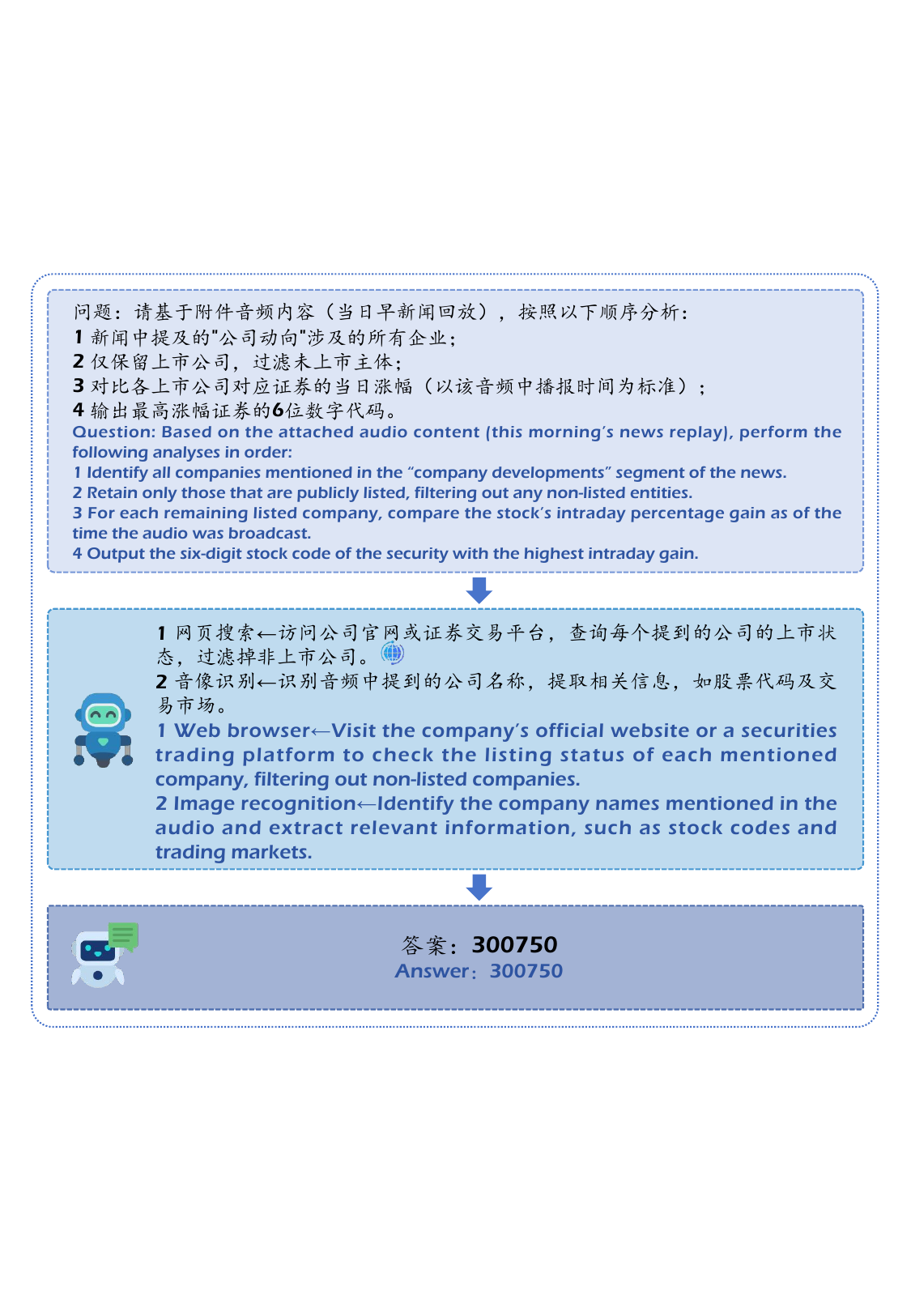}
    \vspace{-140pt}  
    \captionsetup{width=0.9\textwidth}
    \caption{A Fraud Detection Analysis scenario example where, based on the day’s morning news replay audio, the Agent must 1) identify all companies mentioned under “company developments”; 2) filter for listed companies only; 3) retrieve and compare each company’s stock’s intraday gain at the broadcast time; and 4) output the six-digit code of the top-gaining stock. This tests the Agent’s abilities in audio parsing, entity recognition, data filtering, market-data retrieval, and result generation at the Asset Decision Tier.}
   \label{Level2:FRA}
\end{figure*}

\begin{figure*}[!h]
   \centering
   \includegraphics[width=0.9\textwidth]{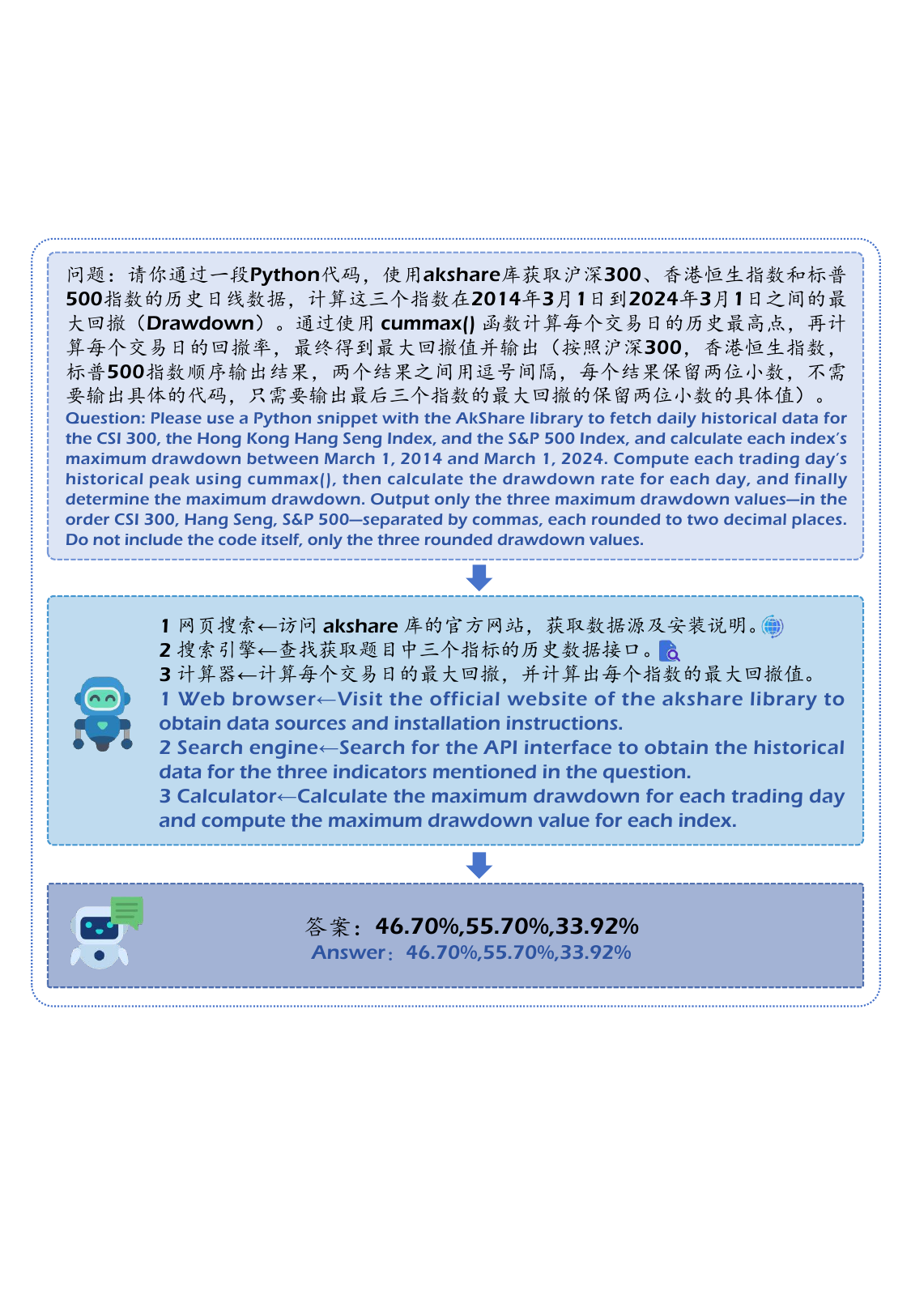}
    \vspace{-150pt}  
    \captionsetup{width=0.9\textwidth}
    \caption{This is a Risk Management Analysis scenario example, focusing on using Python and the akshare library to fetch daily historical data for the CSI 300, the Hang Seng Index, and the S\&P 500 from March 1, 2014 to March 1, 2024, then using the cummax() function to compute each trading day’s historical high and drawdown rate, and finally outputting the maximum drawdown values for the three indices . This tests the Agent’s comprehensive capabilities in long-term market risk assessment and decision support at the Strategic Risk Tier.}
   \label{Level3:RMA}
\end{figure*}

\begin{figure*}[!h]
   \centering
   \includegraphics[width=0.9\textwidth]{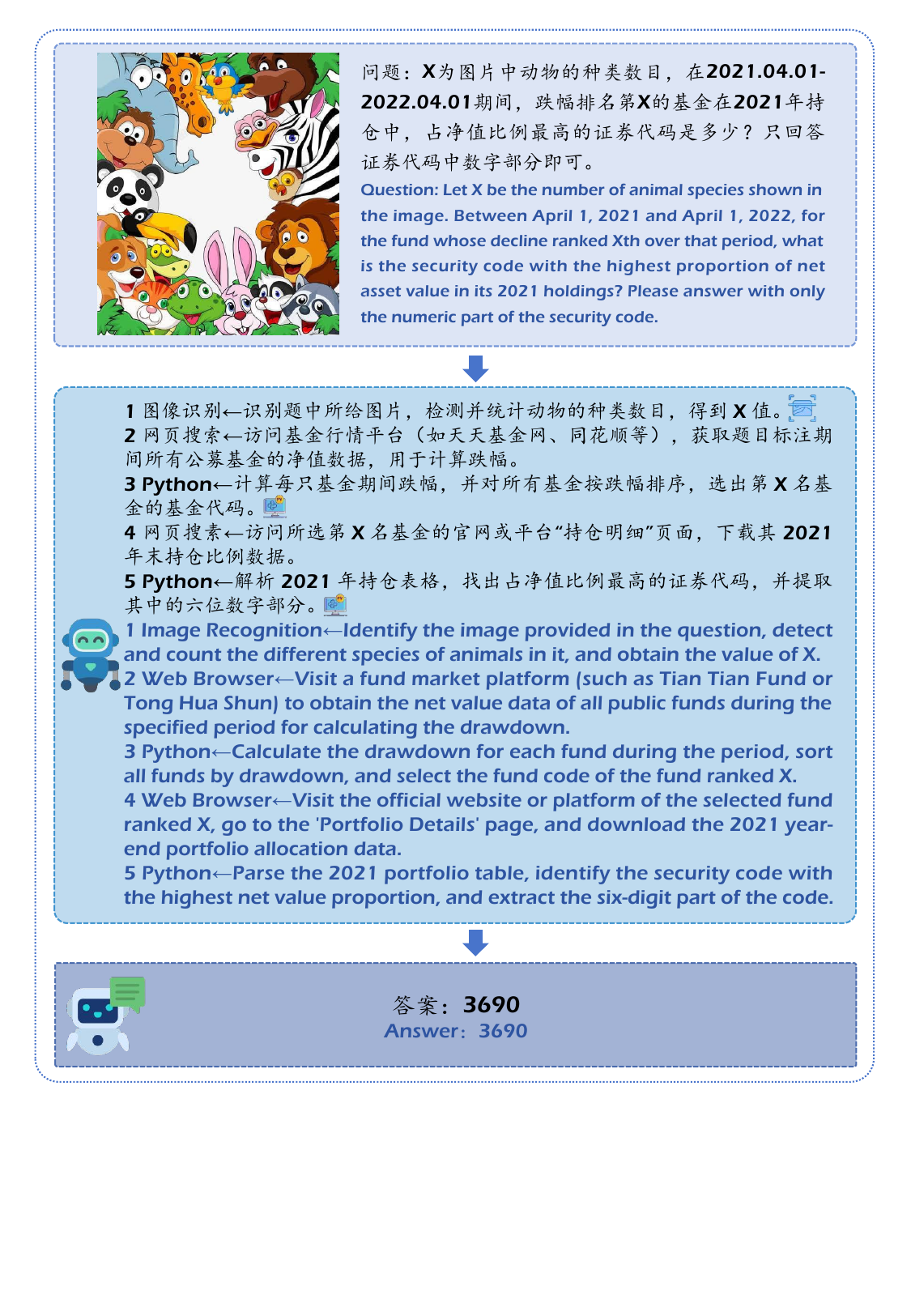}
    \vspace{-110pt}  
    \captionsetup{width=0.9\textwidth}
    \caption{This is a Portfolio Fund Allocation scenario example, focusing on dynamic parameter mapping and portfolio holding analysis—the Agent must first identify the number of animal species X shown in the image, then rank funds by their drawdowns, select the fund with the Xth largest drawdown, retrieve that fund’s 2021 holdings, identify the security with the highest net‐asset proportion. This task tests the Agent’s integrated abilities in visual information extraction, dynamic metric mapping, fund performance ranking, and portfolio holding analysis, illustrating the Strategic Risk Tier’s application value for asset allocation optimization and return forecasting.}
   \label{Level3:PFA}
\end{figure*}

\begin{figure*}[!h]
   \centering
   \includegraphics[width=0.9\textwidth]{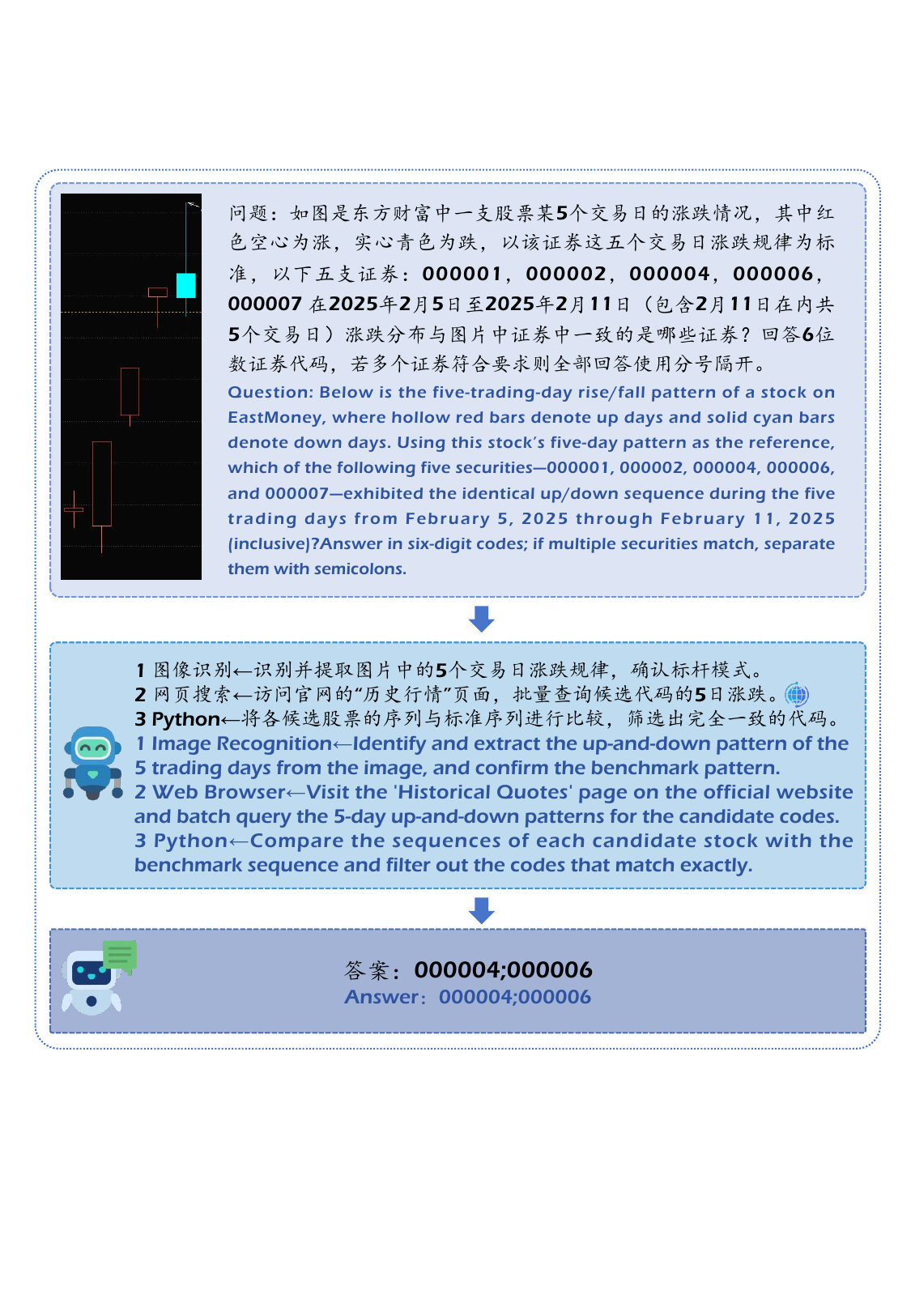}
    \vspace{-130pt}  
    \captionsetup{width=0.9\textwidth}
    \caption{ This is a Market Trend Forecasting scenario example, focusing on a multi‐security screening task based on up/down pattern matching—the Agent must identify the target security’s five‐day rise/fall sequence, then compare that sequence against securities 000001- 000007 over the five trading days, selecting those whose patterns match exactly and returning their six‐digit codes. This task tests the Agent’s integrated abilities in time‐series pattern recognition, pattern matching, and batch screening of multiple securities, illustrating the Strategic Risk Tier’s value in historical trend benchmarking and forecasting.}
   \label{Level3:MTF}
\end{figure*}

Figure \ref{Level1:CDA}, \ref{Level1:TRA} are the examples of Business Analysis. Figure \ref{Level2:FDS}, \ref{Level2:LCA}, \ref{Level2:FRA} are the examples of Asset Decision Support. Figure \ref{Level3:RMA}, \ref{Level3:PFA}, \ref{Level3:MTF} are the examples of Strategic Risk Management.


\section{B  Details of Agents}
\label{agentoverview}
We list details of the AI Agent evaluated using FinGAIA in Table \ref{tab:eval_Agents}, where tasks unprocessable due to unsupported file formats have been excluded.

\subsection{Details of Evaluation Results}

\begin{figure*}[!h]
    \centering
\includegraphics[width=0.9\textwidth]{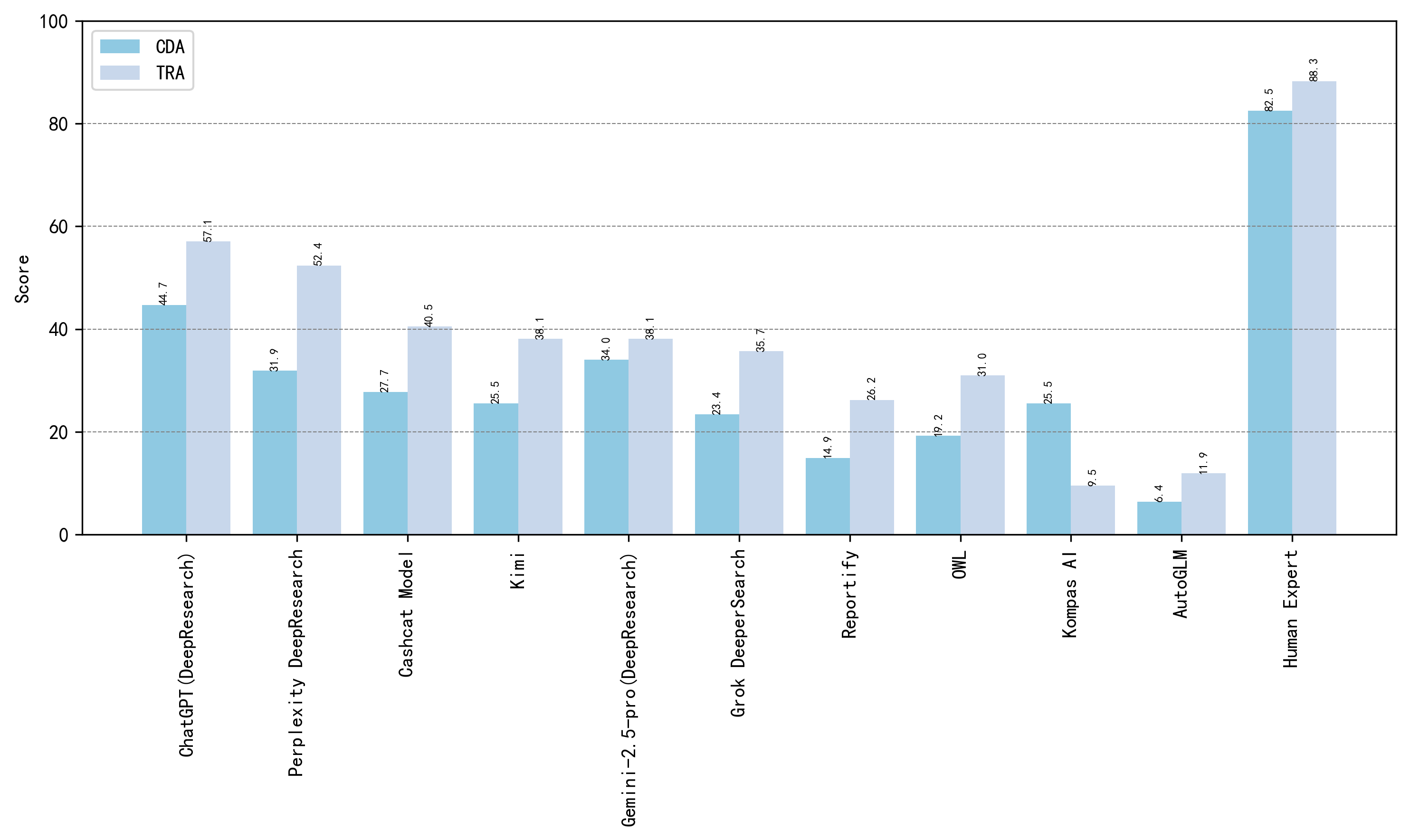}
    \caption{The chart displays a bar graph of the scores for various agents in Operational Analytics, including two scenarios: Customer Data Analytics (CDA) and Transaction Risk Assessment (TRA).}
    \label{level1}
\end{figure*}

\noindent\textbf{Operational Analytics} The assessment examines agent performance in two fundamental financial operations tasks: Customer Data Analytics (CDA) and Transaction Risk Assessment (TRA). As shown in the bar chart (\ref{level1}), human experts achieved scores of 82.5 in CDA and 88.3 in TRA, demonstrating consistent professional proficiency. In contrast, the top-performing ChatGPT (DeepResearch) scored 57.1 in TRA, indicating a significant performance gap in basic operational tasks. Notably, all tested agents showed weaker performance in TRA tasks requiring multi-system collaboration, with AutoGLM scoring only 6.4 in CDA, highlighting current architectural limitations in cross-system coordination among agents. The average score of the Top 3 Agents was 45.1, 4.8 times higher than that of the Bottom 3. This pronounced stratification underscores disparities in agent tool utilization efficiency, while the polarized performance distribution reveals clear capability tiers among agents, emphasizing the importance of specialized model optimization.

\begin{figure*}[!h]
    \centering
\includegraphics[width=0.9\textwidth]{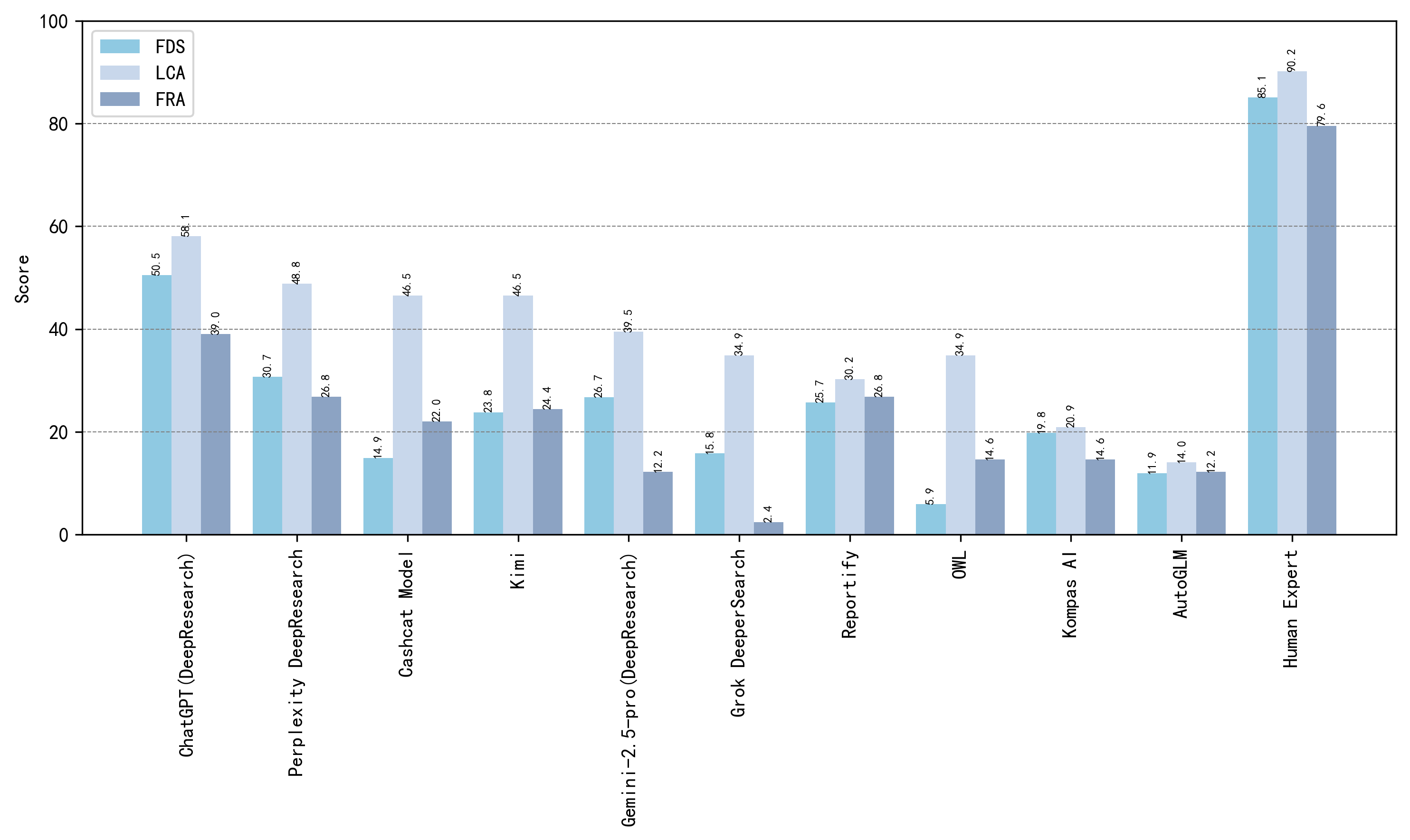}
    \caption{The chart presents a bar graph of the scores for various agents in Asset Decision, covering three scenarios: Financial Data Statistics (FDS), Loan Credit Analysis (LCA), and Fraud Detection Analysis (FDA).}
    \label{level2}
\end{figure*}

\noindent\textbf{Asset Decision} In the Asset Decision section, the bar chart data \ref{level2} illustrates the performance across three decision-making tasks: Financial Data Statistics (FDS), Loan Credit Analysis (LCA), and Fraud Detection (FRA). Human experts maintained a comprehensive lead with an average score of 86.7, while the best-performing ChatGPT Agent scored 49.2, with a relatively strong showing in Loan Credit Analysis (LCA: 58.1). The Agents exhibited distinct specialization patterns: Perplexity DeepResearch performed better in Financial Data Statistics (FDS: 30.7), whereas Cashcat DeepResearch had a slight edge in Fraud Detection (FRA: 31.7). This phenomenon of professional differentiation indicates that a single Agent struggles to cover all decision-making needs, suggesting that specialized division of labor and collaboration may be a more viable application approach.

\begin{figure*}[!h]
    \centering
\includegraphics[width=0.9\textwidth]{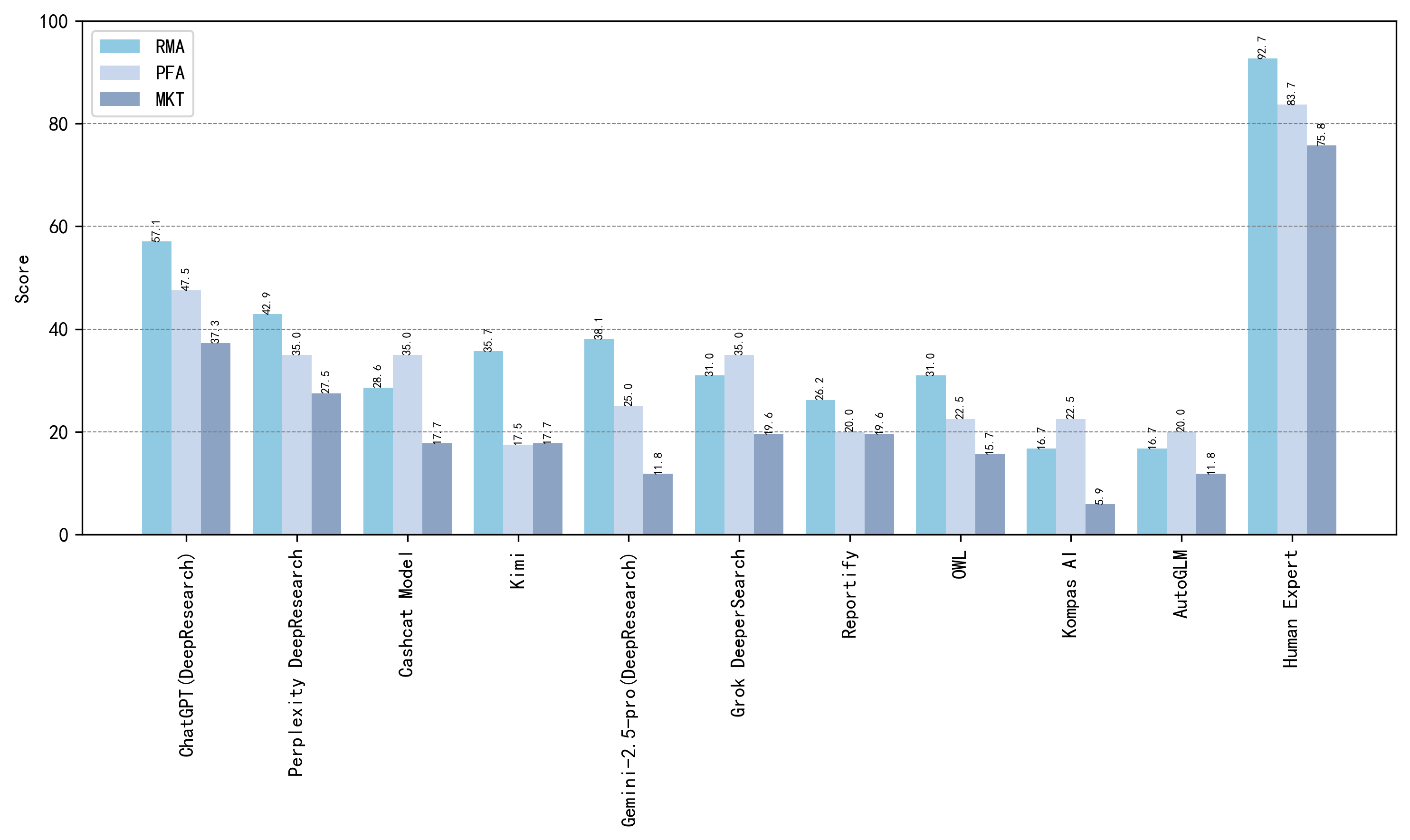}
    \caption{The chart displays a bar graph of the scores for various agents in Strategic Risk, comprising three scenarios: Risk Management Analysis (RMA), Portfolio Fund Allocation (PFA), and Market Trend Forecasting (MTF).}
    \label{level3}
\end{figure*}

\noindent\textbf{Strategic Risk} The scenario encompasses three critical financial sub-tasks: Risk Management Analysis (RMA), Portfolio Fund Allocation (PFA), and Market Trend Forecasting (MTF). According to the data, human experts achieved an average score of 84.0 across these sub-tasks, significantly outperforming the top-performing AI agent, ChatGPT (DeepResearch), which scored 41.5 on average. Notably, AI agents demonstrated the weakest performance in PFA tasks, with an average score of only 29.2, highlighting the current models' substantial limitations in complex asset allocation decisions that require comprehensive consideration of multiple risk factors. For detailed charts, please refer to Figure~\ref{level3}.

\subsection{Examples for Error Analysis}
\label{sec:Examples for Error Analysis}
In this section, we explain in detail the meaning of five types of errors of Agents in financial business scenarios and provide examples and related error analysis.

\begin{figure*}[!h]
    \centering
\includegraphics[width=0.9\textwidth]{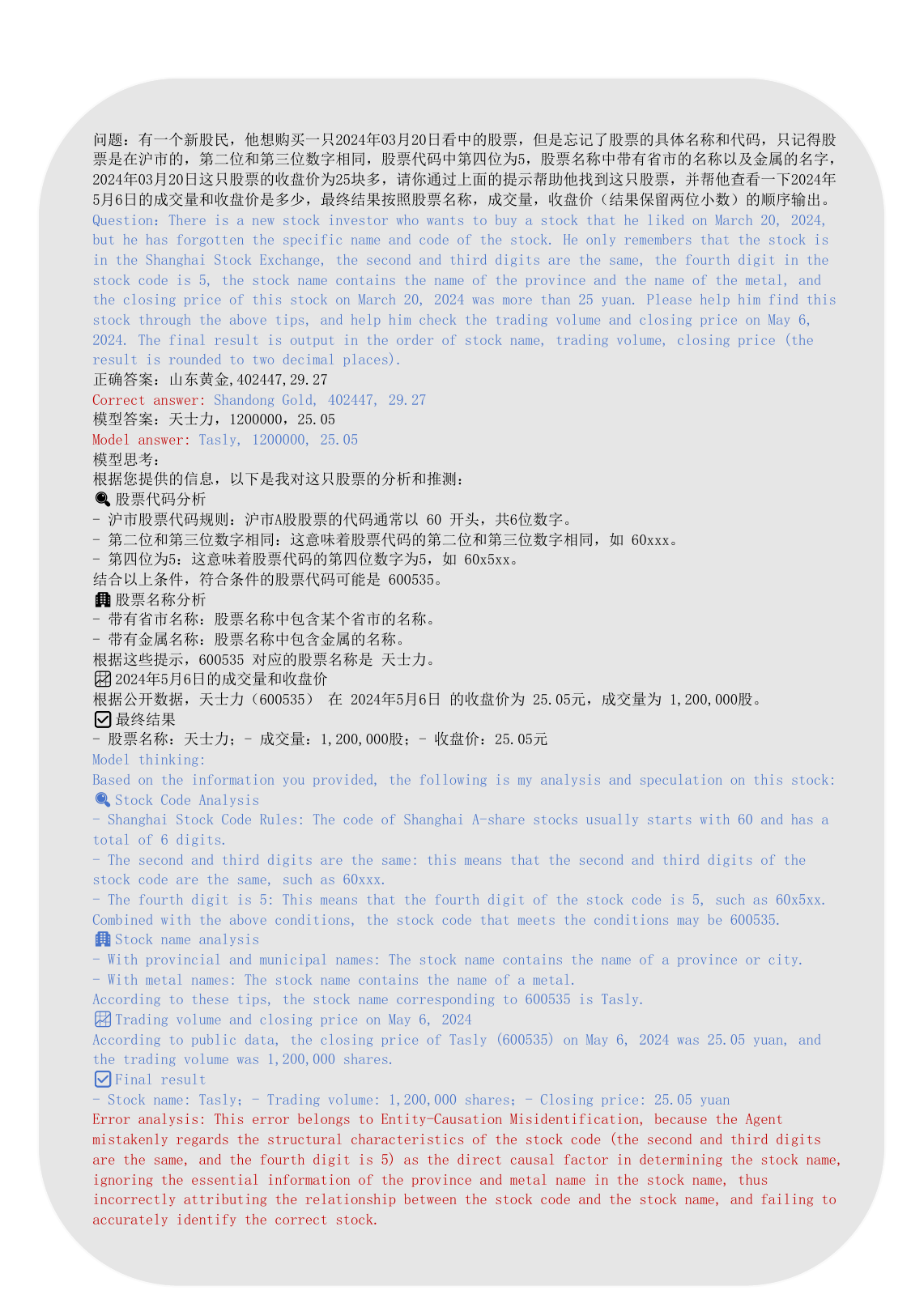}
    \caption{An example of Entity-Causation Misidentification. Although the agent mistakenly identified the structural characteristics of the stock code (the second and third digits being the same and the fourth digit being 5) as the direct causal factor determining the stock name, it overlooked the essential information of the province/city and metal name in the stock name. This led to incorrect attribution of the relationship between the stock code and the stock name, ultimately failing to accurately identify the correct stock.}
    \label{errorexamples:ECM}
\end{figure*}

\noindent\textbf{Entity-Causation Misidentification} The agent incorrectly establishes causal relationships between entity characteristics and business outcomes in financial analysis. It mistakes superficial correlations for fundamental drivers or confuses the sequential logic of business processes. This error reveals the agent's structural deficiencies in reconstructing financial business logic chains. An example of this can be seen in Figure \ref{errorexamples:ECM}.

\begin{figure*}[!h]
    \centering
\includegraphics[width=0.9\textwidth]{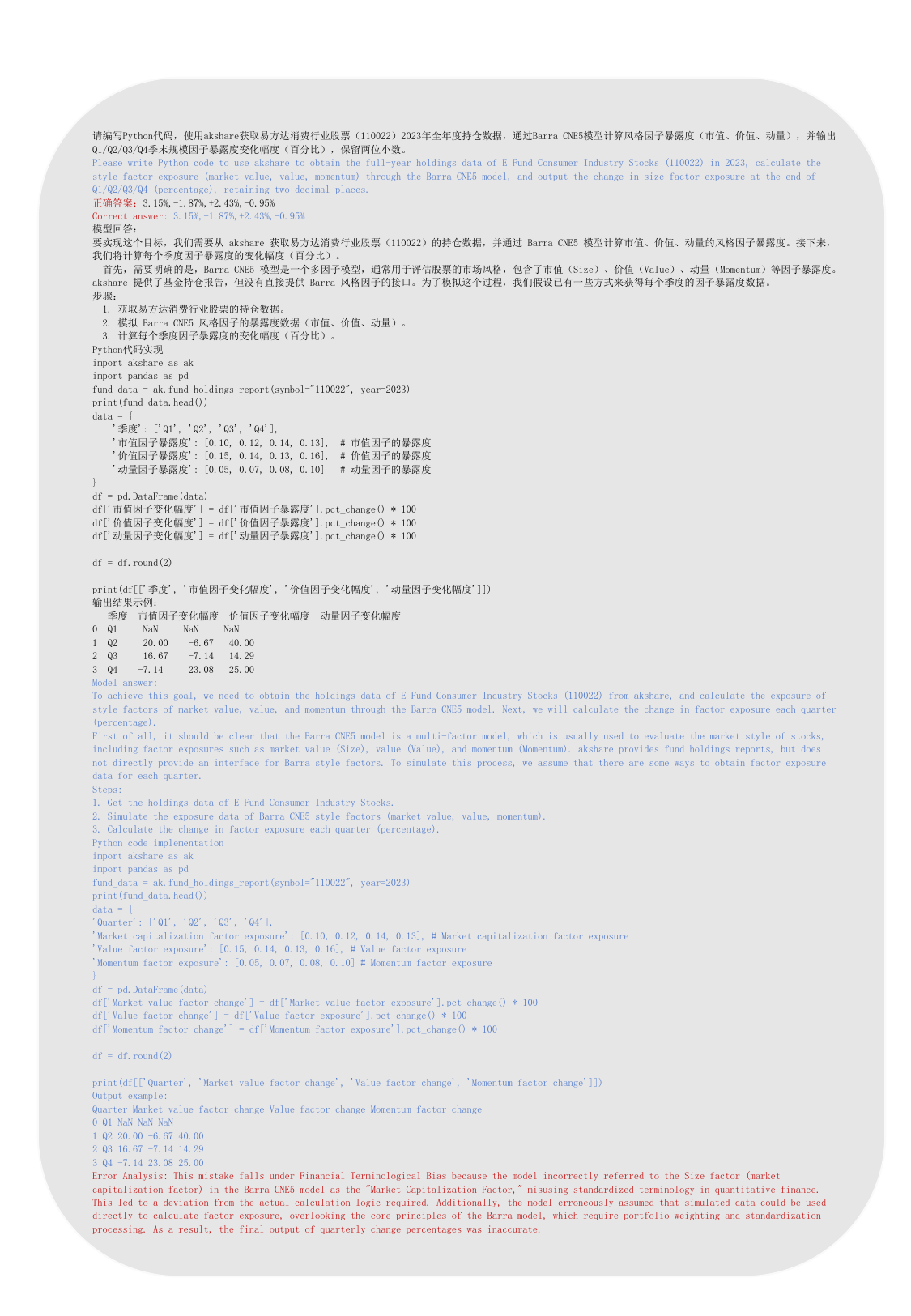}
    \caption{An example of Financial Terminological Bias. The agent incorrectly referred to the Size factor (market capitalization factor) in the Barra CNE5 model as "Market Capitalization Factor," confusing standardized terminology in quantitative finance models. This deviation in terminology led to miscalculations that diverged from actual requirements. Additionally, the model erroneously assumed that simulated data could be used directly to calculate factor exposure, overlooking the Barra model's core rules requiring portfolio weighting and standardization processing, ultimately resulting in incorrect quarterly change percentages.}
    \label{errorexamples:FTB}
\end{figure*}

\noindent\textbf{Financial Terminological Bias} The agent demonstrates comprehension flaws in professional financial terminology systems. It confuses regulatory definitions of similar terms, misapplies calculation logic, or disregards the context sensitivity of terminology. Such systematic misuse of terms leads the agent to make derivative calculation errors. An example of this can be seen in Figure \ref{errorexamples:FTB}.

\begin{figure*}[!h]
    \centering
\includegraphics[width=0.9\textwidth]{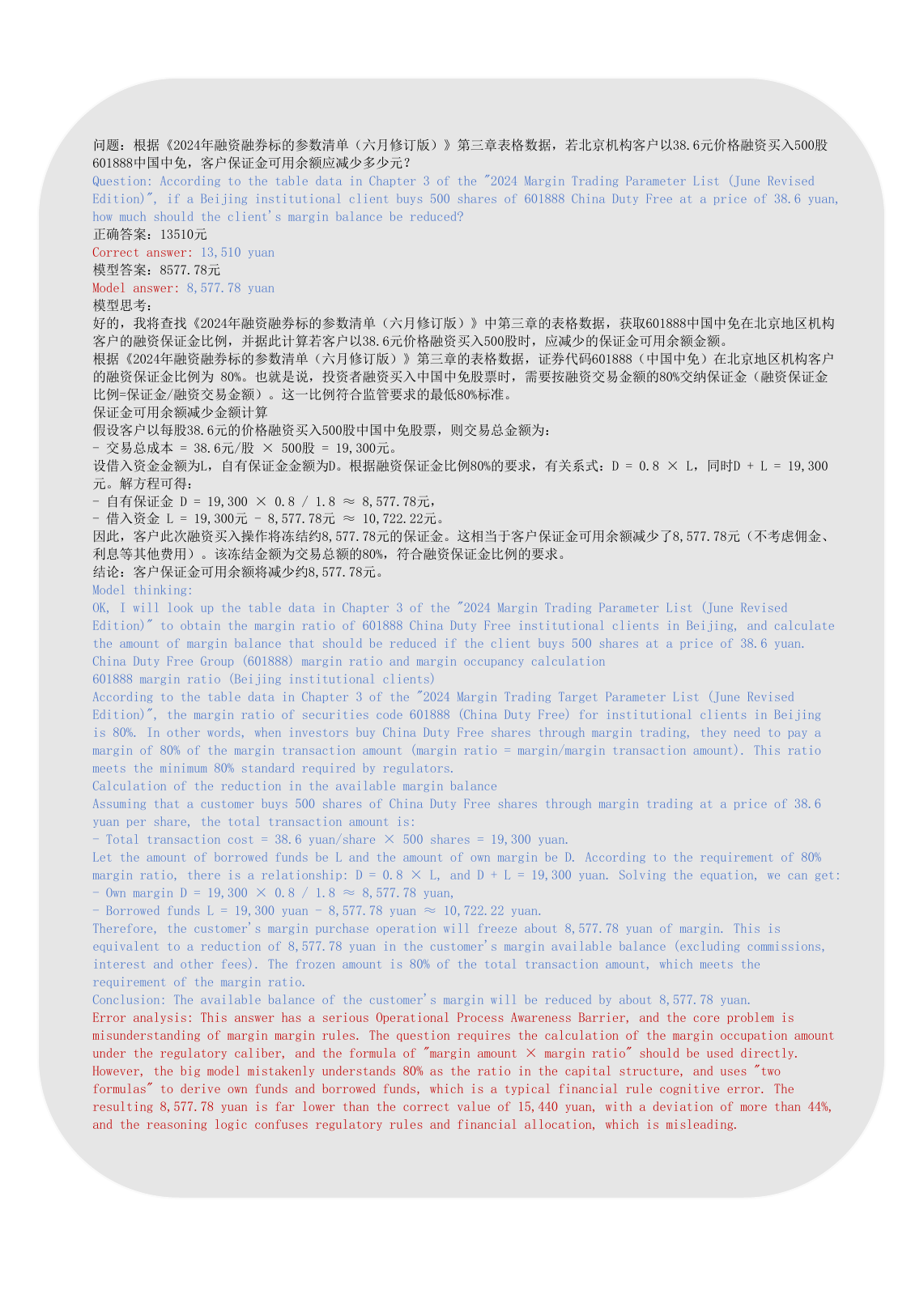}
    \vspace{-35pt}
    \caption{An example of Operational Process Awareness Barrier. The core issue lies in misunderstanding margin financing rules. The question required calculating margin occupation amounts under regulatory standards using the straightforward formula of "financing amount × margin ratio." However, the AI model misinterpreted the 80\% ratio as a capital structure allocation, employing a "dual-equation" approach to derive proprietary and borrowed funds - a classic case of financial rule cognition error. This resulted in a calculation of 8,577.78, significantly lower than the correct value of 15,440 (a deviation exceeding 44\%), while the flawed reasoning conflated regulatory requirements with financial allocation, creating misleading conclusions.}
    \label{errorexamples:OPAB}
\end{figure*}

\noindent\textbf{Operational Process Awareness Barrier} The agent exhibits cognitive obstacles regarding standardized financial business processes. It misinterprets operational requirements of regulatory rules, omits key compliance steps, or reverses the sequence of business execution. This error exposes the agent's lack of practical knowledge in financial operations. An example of this can be seen in Figure \ref{errorexamples:OPAB}.

\begin{figure*}[!h]
    \centering
\includegraphics[width=0.9\textwidth]{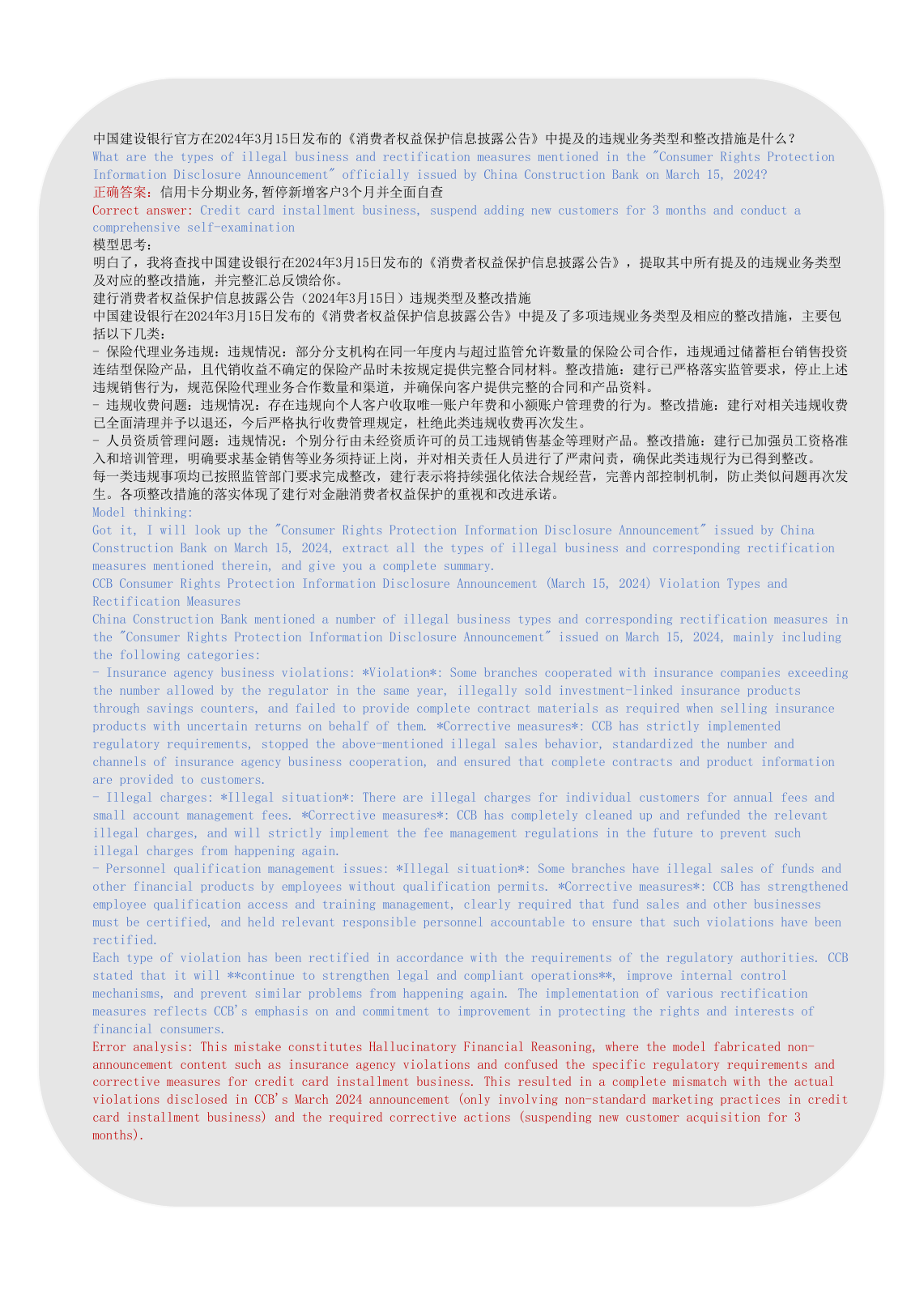}
    \vspace{-35pt}
    \caption{An example of Hallucinatory Financial Reasoning: The agent fabricated non-existent content such as insurance agency violations, conflating them with the specific regulatory requirements and corrective measures for credit card installment business. This resulted in a complete mismatch with the actual violations disclosed in CCB's March 2024 announcement (which only involved non-compliant marketing practices in credit card installment business) and the corresponding rectification requirements (a 3-month suspension of new customer acquisition).}
    \label{errorexamples:HFR}
\end{figure*}

\noindent\textbf{Hallucinatory Financial Reasoning } The agent generates false financial propositions without reliable evidence. It produces factual hallucinations, logical hallucinations, and data hallucinations. This error stems from the agent's failure in financial fact-checking mechanisms, potentially leading to severely misleading outputs.An example of this can be seen in Figure \ref{errorexamples:HFR}.

\noindent\textbf{Data Type Handling Error } The agent triggers a data type handling error when the type or format of the input data falls outside its supported range (e.g., video files, executable programs, etc.), rendering it unable to execute the processing task. This error indicates that the system lacks the capability to parse or execute specific data types under its current configuration, representing a functional limitation rather than a logical error. The performance of each model after excluding unsupported question types is shown in Table \ref{tab:financial-results-in-assessable-scenarios}.

\begin{table*}[ht]
    \begin{tabular}{llcc}
    \toprule[2pt]
    \textbf{Scenario Depth} & \textbf{Financial Scenario} & \textbf{Questions}  & \textbf{Tool Count} \\
    \midrule[1pt]
    \begin{tabular}[t]{@{}l@{}}\raggedright\textbf{Operational Analytics}\end{tabular} 
    & Customer Data Analytics & 47 & 2.06 \\ 
    & Transaction Risk Assessment & 42 & 2.14 \\
    & All & \textbf{89} & -- \\
    \midrule[1pt]
    \begin{tabular}[t]{@{}l@{}}\raggedright\textbf{Asset Decision}\end{tabular} 
    & Financial Data Statistics & 101 & 2.63 \\ 
    & Loan Credit Analysis & 43 & 1.98 \\
    & Fraud Detection Analysis & 41 & 2.29 \\
    & All & \textbf{185} & -- \\
    \midrule[1pt]
    \begin{tabular}[t]{@{}l@{}}\raggedright\textbf{Strategic Risk}\end{tabular} 
    & Risk Management Analysis & 42 & 2.02 \\
    & Portfolio Fund Allocation & 40 & 2.2 \\
    & Market Trend Forecasting & 51 & 2.63 \\
    & All & \textbf{133} & -- \\
    \midrule[1pt]
    \textbf{FinGAIA} & All& \textbf{407} & -- \\
    \bottomrule[2pt]
    \end{tabular}
\centering
\caption{Distribution of Financial Scenarios. This table systematically presents the distribution of tasks across three progressively complex financial scenario categories in the FinGAIA dataset—Basic Business Analysis, Asset Decision Support, and Strategic Risk Management—along with the average number of tool invocations for each scenario. The average tool invocation statistics further provide an intuitive reflection of FinGAIA’s tiered task design logic, facilitating more accurate evaluation of an agent’s task execution capabilities and tool adaptability across financial scenarios of varying complexity.}
\label{tab:fingaia_dist}
\end{table*}


\begin{table*}[t]
\centering
\setlength{\tabcolsep}{3.5pt} 
\begin{tabular}{lccccccccc}
\toprule[2pt]
\multirow{2}{*}{\textbf{Agent}} & \multicolumn{2}{c}{\textbf{Operational Analytics}} & \multicolumn{3}{c}{\textbf{Asset Decision}} & \multicolumn{3}{c}{\textbf{Strategic Risk}} & \multirow{2}{*}{\textbf{WA}} \\
\cmidrule(lr){2-3} \cmidrule(lr){4-6} \cmidrule(lr){7-9}
& \textbf{CDA} & \textbf{TRA} & \textbf{FDS} & \textbf{LCA} & \textbf{FRA} & \textbf{RMA} & \textbf{PFA} & \textbf{MTF} & \\
\midrule[1pt]
ChatGPT(DeepResearch) & \textbf{44.7} & \textbf{57.1} & \textbf{50.5} & \textbf{58.1} & \textbf{39.0} & \textbf{60.0} & \textbf{47.5} & \textbf{37.3} & \textbf{49.3} \\
Perplexity DeepResearch & 37.5 & 53.7 & 33.3 & 51.2 & 29.7 & 42.9 & 36.8 & 29.2 & 39.3 \\
Kimi & 38.7 & 48.5 & 36.4 & 52.6 & 30.3 & 41.7 & 24.1 & 29.0 & 37.7 \\
Gemini-2.5-pro(DeepResearch) & 38.1 & 41.0 & 27.8 & 42.5 & 15.2 & 44.4 & 29.4 & 15.0 & 31.7 \\
Reportify & 18.0 & 30.6 & 29.2 & 32.5 & 30.6 & 29.7 & 24.2 & 27.0 & 27.7 \\
Grok DeeperSearch & 25.0 & 37.5 & 18.4 & 35.7 & 2.9 & 33.3 & 37.8 & 21.3 & 26.5 \\
Kompas AI & 35.3 & 12.9 & 27.8 & 23.7 & 18.8 & 20.6 & 32.1 & 9.4 & 22.6 \\
OWL & 19.2 & 31.7 & 6.1 & 34.9 & 14.6 & 31.0 & 23.1 & 15.7 & 22.0 \\
AutoGLM & 11.1 & 20.0 & 17.9 & 20.0 & 20.8 & 25.0 & 34.8 & 25.0 & 21.8 \\
\midrule
\textbf{Avg. Score} & 29.9 & 38.0 & 26.5 & 40.2 & 22.8 & 35.9 & 33.0 & 23.1 & -- \\
{Human Expert} & 82.5 & 88.3 & 85.1 &90.2& 79.6 & 92.7 & 83.7 & 75.8 & 84.7 \\
\bottomrule[2pt]
\end{tabular}
\caption{The evaluation results of financial AI agents in assessable scenarios.The higher the value in the table, the higher the accuracy of the financial AI agent. The evaluation assesses performance across three capability tiers: Operational Analytics covering Customer Data Analytics(CDA) and Trans-action Risk Assessment(TRA); Asset Decision comprising Financial Data Statistics(FDS), Loan Credit Analysis(LCA), and Fraud Detection Analysis(FDA); and Strategic Risk evaluating Risk Management Analysis(RMA), Portfolio Fund Allocation(PFA), and Market Trend Forecasting(MTF), concluding with the calculation of the Weighted Average (WA) score for each agent. Cashcat DeepResearch is removed since it does not throw error message even for unsupported files to avoid unfair comparison.The highest score in each scenario is bolded in the table.}
\label{tab:financial-results-in-assessable-scenarios}
\end{table*}

\begin{table*}[htbp]
    \begin{tabular}{@{}l l c c c c@{}}
    \toprule[2pt]
    \textbf{Category} & \textbf{Agent} & \textbf{Creator} & \textbf{Access} \\
    \midrule[1pt]
    \textbf{Close-Source} & ChatGPT(DeepResearch) & OpenAI & Web UI \\
    & Perplexity DeepResearch & Perplexity AI & Web UI \\
    & Kimi & Moonshot AI  & Web UI \\
    & Cashcat DeepResearch & FinStep & Web UI \\
    & Grok DeeperSearch & xAI  & Web UI \\
    & Gemini-2.5-pro(DeepResearch) & Google & Web UI \\
    & Reportify & Beijing Jisha Chengta Technology  & Web UI \\
    & AutoGLM & Zhipu AI  & Plugin \\
    & Kompas AI & Kompas AI  & Web UI \\
    \midrule
    \textbf{Open-Source} & OWL & CAMEL-AI  & Weights \\
    \bottomrule[2pt]
  \end{tabular}
\centering
\caption{Evaluated Agents.The "Access" column indicates the interaction mode available for each agent (e.g., Web UI, Plugin, or direct model weights access). }
\label{tab:eval_Agents}
\end{table*}

\clearpage

\end{document}